\documentclass[letterpaper]{article} 

\usepackage{aaai24}  
\usepackage{times}  
\usepackage{helvet}  
\usepackage{courier}  
\usepackage[hyphens]{url}  
\usepackage{graphicx} 
\usepackage{natbib}  
\usepackage{caption} 
\usepackage{algorithm}
\usepackage{algorithmic}
\usepackage{newfloat}
\usepackage{listings}
\DeclareCaptionStyle{ruled}{labelfont=normalfont,labelsep=colon,strut=off} 
\floatstyle{ruled}
\newfloat{listing}{tb}{lst}{}
\floatname{listing}{Listing}
\usepackage{graphicx}
\usepackage[dvipsnames]{xcolor}

\usepackage{amsmath}
\usepackage[capitalize,nosort]{cleveref}
\crefname{section}{Sec.}{Secs.}
\Crefname{section}{Section}{Sections}
\Crefname{table}{Table}{Tables}
\crefname{table}{Tab.}{Tabs.}
\Crefname{appsec}{Appendix}{Appendices}
\crefname{appsec}{App.}{Apps.}

\usepackage{makecell}
\usepackage{multirow}
\usepackage{pifont}
\usepackage{booktabs}
\usepackage{caption}
\usepackage{subcaption}
\usepackage{amsfonts}

\newcommand{\cmark}{\ding{51}}
\newcommand{\xmark}{\ding{55}}

\DeclareMathOperator*{\argmax}{arg\,max}

\usepackage{amsthm}

\theoremstyle{definition}
\newtheorem{definition}{Definition}[section]

\title{Interpretability Benchmark for Evaluating Spatial Misalignment\\of Prototypical Parts Explanations}

\author{
    Mikołaj Sacha\textsuperscript{\rm 1, 2},
    Bartosz Jura\textsuperscript{\rm 3},
    Dawid Rymarczyk\textsuperscript{\rm 1,2}, \\
    Łukasz Struski\textsuperscript{\rm 1},
    Jacek Tabor\textsuperscript{\rm 1},
    Bartosz Zieliński\textsuperscript{\rm 1,4}
}
\affiliations{

    \textsuperscript{\rm 1}Faculty of Mathematics and Computer Science, Jagiellonian University\\
    \textsuperscript{\rm 2}Doctoral School of Exact and Natural Sciences, Jagiellonian University\\
    \textsuperscript{\rm 3} Faculty of Management and Social Communication, Jagiellonian University\\
    \textsuperscript{\rm 4}IDEAS NCBR\\
    mikolaj.sacha@doctoral.uj.edu.pl
}

\begin{document}

\maketitle

\begin{abstract}
Prototypical parts-based networks are becoming increasingly popular due to their faithful self-explanations. However, their similarity maps are calculated in the penultimate network layer. Therefore, the receptive field of the prototype activation region often depends on parts of the image outside this region, which can lead to misleading interpretations. We name this undesired behavior a spatial explanation misalignment and introduce an interpretability benchmark with a set of dedicated metrics for quantifying this phenomenon. In addition, we propose a method for misalignment compensation and apply it to existing state-of-the-art models. We show the expressiveness of our benchmark and the effectiveness of the proposed compensation methodology through extensive empirical studies.
\end{abstract}

\maketitle


%

\section{Introduction}


The lack of insights into the reasons behind model predictions is a major limitation of current deep learning-based systems, particularly in high-stake decision fields like medicine and autonomous driving~\cite{rudin2019stop}. As a result, eXplainable Artificial Intelligence (XAI) has gained significant attention in recent years, with two main branches of research being extensively developed: post hoc and self-explainable methods~\cite{rudin2019stop}.

The post hoc approaches assume that an explainer model needs to be developed to explain the predictions of a classic deep neural network. However, this approach may be biased and unreliable~\cite{adebayo2018sanity}. That is why self-explainable methods were introduced, such as prototypical parts-based methods~\cite{chen2019looks}. They contain built-in interpretability components and provide interpretation along with the prediction.

\begin{figure}[t]
  \centering
  \includegraphics[width=0.85\columnwidth]{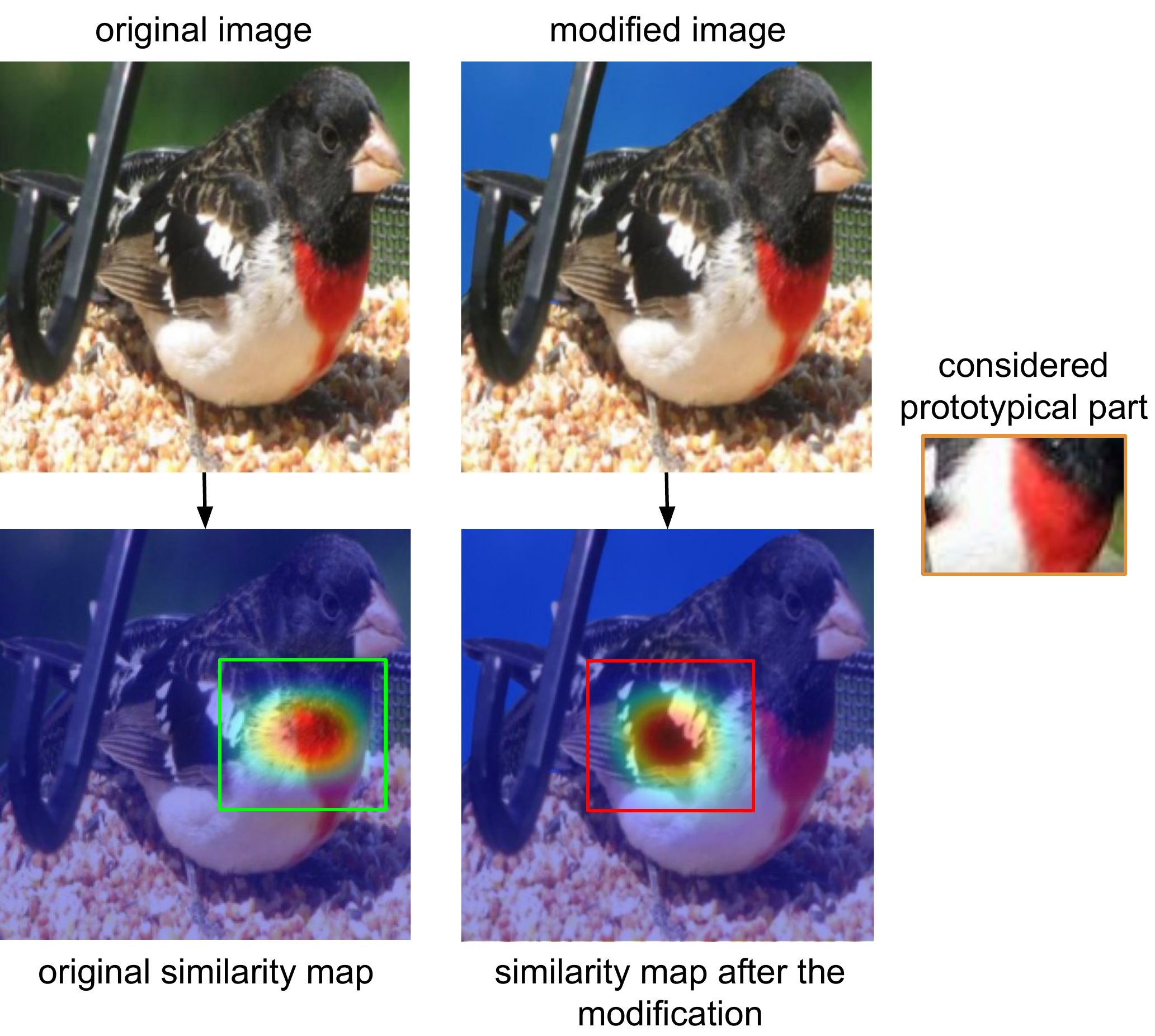}
  \caption{The receptive field of the prototypical part activation region often depends on parts of the image outside this region. In this example, the activation region (green bounding box) depends on the upper left background outside the bounding box. Therefore, after changing this part of the background (as in the modified image), the location of the activation region also changes (red bounding box). Such behavior is unwanted because it misleads explanations.}
  \label{fig:new_teaser}
  \vspace{-1em}
\end{figure}

Prototypical parts-based networks, such as ProtoPNet~\cite{chen2019looks}, utilize feature-matching learning theory~\cite{rosch1975cognitive} to identify important image parts by comparing them with reference patterns from training data. However, despite its ability to provide highly faithful explanations, this approach has known shortcomings, such as ambiguity of prototypical parts~\cite{nauta2020looks} or non-resistance to image modifications, like JPEG compression~\cite{hoffmann2021looks}. Consequently, tools such as PRP~\cite{gautam2023looks} have been introduced to improve interpretability.

In this paper, we identify another risk related to the fact that prototypical parts similarity maps are calculated in the penultimate network layer. Therefore, the receptive field of the prototypical part activation region often depends on parts of the image outside this region. It can result in misleading explanations because users usually identify the activation region with the receptive field, while, as presented in \cref{fig:new_teaser}, this assumption is only sometimes fulfilled.

To assess the explanation misalignment of prototypical parts-based methods, we introduce an interpretability benchmark with a set of dedicated metrics. It adversarially modifies the input image to reduce the high activation of a prototypical part by changing only the image area where a particular prototypical part is almost inactive. We decided to use adversarial modification due to its flexibility which allows us to choose the modified pixels and strength of the modification without altering the input image too much. We use the original and modified images to compute easy-to-interpret explanation misalignment metrics.

Except for the interpretability benchmark, we propose a novel compensation methodology preserving the spatial relationship between the prototypical part activation region and its receptive field. It is based on a novel loss function computed for the image passed twice through the network, with and without a mask, and masking-based augmentation.

Finally, we provide an extensive experimental evaluation of our compensation methodology using state-of-the-art prototypical parts methods, showing the effectiveness of our benchmark and the constructed compensation method.

Our contributions can be summarized as follows:
\begin{itemize}
    \item We systematize the limitations of the basic visualization of prototypical parts activation, which can lead to misleading explanations.
    \item We propose an interpretability benchmark for measuring the spatial misalignment of the prototypical part activation region and its receptive field to assess the reliability of explanations.
    \item We introduce a novel compensation methodology for this misalignment that can be used with any prototypical parts-based model.
\end{itemize}

\section{Related works}

Methods used to explain deep learning models can be classified into post hoc and self-explainable~\cite{rudin2019stop}. Post hoc methods assume that the reasoning process is hidden within a black box model, and a new explainer model must be created to reveal it. Some of those methods generate saliency maps~\cite{rebuffi2020there,selvaraju2019taking,simonyan2014deep} or use Concept Activation Vectors (CAV) to construct explanation with user-friendly concepts~\cite{chen2020concept,NEURIPS2019_77d2afcb,kim2018interpretability,NEURIPS2020_ecb287ff}. Others provide counterfactual examples~\cite{abbasnejad2020counterfactual,niu2021counterfactual} or analyze the network's reaction to image perturbations~\cite{basaj2021explaining,fong2019understanding,ribeiro2016should}. The post hoc methods are easy to implement as they do not interfere with the architecture. However, they may produce biased and unreliable explanations~\cite{NEURIPS2018_294a8ed2}. Therefore, considerable effort has been devoted to designing self-explainable models~\cite{NEURIPS2018_3e9f0fc9,brendel2018approximating} that make the internal decision process visible for the user. Many interpretable solutions use attention mechanisms~\cite{liu2021visual,zheng2017learning,zheng2019looking} or exploit the activation space~\cite{Guidotti_Monreale_Matwin_Pedreschi_2020,puyol2020interpretable}, such as adversarial autoencoders. However, the most recent approaches are built on an interpretable method introduced in~\cite{chen2019looks} (ProtoPNet) using a hidden layer of prototypical parts to discover visual concepts.

Multiple self-explainable methods enhance ProtoPNet~\cite{chen2019looks}. TesNet~\cite{wang2021interpretable} constructs the latent space on a Grassman manifold. PIP-Net~\cite{nauta2023pip} redefines the prototypical parts layer to allow out-of-distribution data detection. ProtoVAE~\cite{gautam2022protovae} leverages a variational autoencoder with prototypical parts. ProtoPShare~\cite{rymarczyk2021protopshare}, ProtoTree~\cite{nauta2021neural}, ProtoKNN~\cite{ukailooks}, and ProtoPool~\cite{rymarczyk2022interpretable} reduce the number of prototypical parts used in the classification. ProtoPShare introduces data-dependent merge-pruning that discovers prototypical parts of similar semantics and joins them. ProtoTree uses a soft neural decision tree that may depend on the negative reasoning process and is extended to a visual transformer by~\cite{kim2022transform}. ProtoPool proposes differentiable prototypical parts to class assignments while ProtoKNN adapts distance-based classifiers to prototypical parts. At the same time, more alternative approaches organize the prototypical parts hierarchically~\cite{hase2019interpretable} to classify input at every level of a predefined taxonomy or transform prototypical parts from the latent space to data space~\cite{li2018deep}. Moreover, prototype-based solutions are widely adopted in various fields such as medical imaging~\cite{afnan2021interpretable,barnett2021iaia,kim2021xprotonet,rymarczyk2023protomil,singh2021these}, time-series analysis~\cite{gee2019explaining}, graphs analysis~\cite{rymarczyk2023progrest,zhang2021protgnn}, semantic segmentation~\cite{sacha2023protoseg}, deepfake detection~\cite{trinh2021fake}, zero-shot learning~\cite{Xu2020zeroshot}, and continual learning~\cite{rymarczyk2023icicle}.

As the number of published prototypical parts-based methods grows, the community starts to contemplate the correct ways of comparing them, using not only accuracy. For example, \cite{hoffmann2021looks} investigates ProtoPNets interpretability and discovers a semantic gap between similarity in input and latent space. At the same time, \cite{etmann2019connection,zhang2019interpreting,tsipras2018robustness} highlight connections between the explainability of machine learning models and their adversarial robustness. However, according to our knowledge, no systematic benchmark has been proposed for a comprehensive comparison of prototypical parts-based models, such as the one we propose.

\section{Preliminaries}

\subsection{Prototypical parts network}

To make this work self-contained, in this section, we describe the ProtoPNet model~\cite{chen2019looks}, which introduces the prototypical parts layer. The following paragraphs include the architecture, inference, and basic visualization. We provide the ProtoPNets' training schema in Supplement.

\paragraph{Architecture.}
Prototypical parts networks~\cite{chen2019looks} consist of a backbone convolutional network $f$, a prototypical part layer $g$, and a fully connected layer $h$. The prototypical part layer $g$ consists of $K$ prototypical parts $p \in \mathbb{R}^D$ per class, whose assignment is coded in the fully connected layer $h$. If the prototypical part $p$ is assigned to class $c$, then the weight between them equals $1$. Otherwise, it is set to $-0.5$. We will denote the set of all prototypical parts as $P$ and the set of prototypical parts of class $c$ as $P_c$. 

\paragraph{Inference.}
Given an input image $x$, its representation $f(x)$ of shape $H\times W\times D$ is generated with a backbone $f$. The $H$ and $W$ represent height and width after the last convolutional layer, where $D$ is its depth. Then, each prototypical part $p$ is compared to each of $H\times W$ representation vectors $z_j \in f(x)$ to find the maximum similarity (i.e. the maximal activation of this prototypical part on the input image)
\begin{equation}
\vspace{-0.5ex}
g_{p}(x) = \max_{z_j\in f(x)} sim(p, z_j), 
\vspace{-0.5ex}
\end{equation}
where
\begin{equation}
\vspace{-0.25ex}
   sim(p, z_j)=\log\frac{|z_j-p|_2 + 1}{|z_j-p|_2 + \eta}
\label{eq:sim}
\end{equation}
and $\eta \ll 1$. Suppose $s$ of dimension $W \times H$ refers to the similarity map generated for the whole $f(x)$ representation
\begin{equation}
\vspace{-0.5ex}
   s = sim(p, f(x)).
\label{eq:s}
\end{equation}
In that case, the final prediction is obtained by pushing similarity values through the fully connected layer $h$.


\paragraph{Visualization.}
Visualization of the regions corresponding to prototypical parts is obtained with similarity maps calculated by the layer $g$ before max pooling, upscaled from $H \times W$ to the resolution of the input image, and overlayed. To further simplify the visualization, the authors of ProtoPNet~\cite{chen2019looks} take the $90$th percentile of this upscaled similarity map and draw a bounding box around the highest activation values to mark the prototypical part.

\subsection{Spatial misalignment of explanations}



\begin{definition}
Let us consider a similarity map $s_i=sim(p_i, f(x))$ of the prototypical part $p_i$ for an input image $x$ (as defined in~\cref{eq:s}). Moreover, let $m_i$ be a binarized mask obtained by interpolating $s_i$ to the input resolution and assigning a positive mask value to pixels with an activation value above the $90$th percentile. We define the spatial misalignment of explanations as:
\begin{equation}
\vspace{-0.5ex}
\Delta=\max_{\bar{x}_i} \| sim(p_i, f(x)) - sim(p_i, f(\bar{x}_i)) \|,
\vspace{-0.5ex}
\end{equation}
where $\bar{x}_i$ correspond to any $x$ modification outside mask $m_i$.   
\end{definition}

Observe that if $\Delta=0$, then changes outside the mask do not impact the explanation. Therefore, they are spatially aligned. However, the higher $\Delta$, the larger the misalignment.


\begin{figure}[t]
  \centering
  \includegraphics[width=0.85\columnwidth]{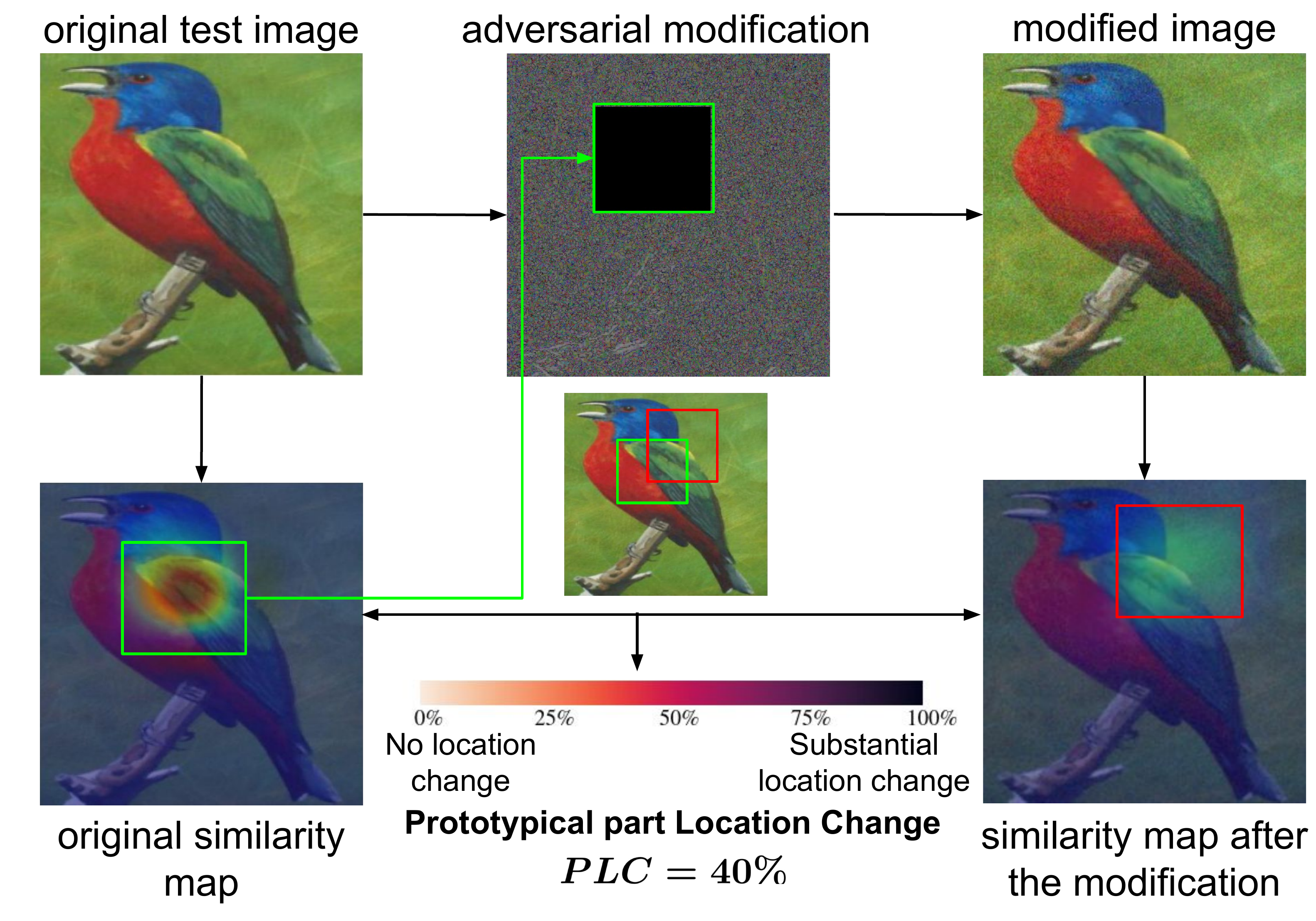}
  \caption{\textit{Prototypical part Location Change} ($PLC$), similarly to the remaining metrics, is a two-step process. The first step is similar for all metrics. It calculates the similarity map for the maximally activated prototypical part and adversarially modifies the image outside the activated region (outside the green bounding box). As a result, the activation region can change (red bounding box). The second step differs between metrics. In the case of $PLC$, it measures the location change of activation before and after modification (difference between green and red bounding box, respectively).}
  \label{fig:teaser}
  \vspace{-1em}
\end{figure}

\section{Interpretability benchmark}
\label{sec:benchmark}
This section introduces a benchmark for evaluating the spatial misalignment of prototypical part explanations. It consists of three metrics that can be considered complementary to the performance metrics by the community researchers.

The central idea is to analyze differences in prototypical part activations (similarity maps) obtained for the original and adversarially modified image. Modifications are made only outside the activation region of the original image. Therefore, in the case of perfect alignment, there should be no difference between prototypical part activations obtained for both images. Otherwise, there can be differences in explanation location, activation, and ranking, considered in the following metrics.

\paragraph{Adversarial modification.}
To formally describe our benchmark, let us assume that $X$ is a test set of images used to quantify the spatial misalignment of the model. Each $x \in X$ is passed through the backbone convolutional network $f$ and compared to all prototypical parts using~\cref{eq:sim}.

Let $\bar{p}$ be a prototypical part with the largest activation on $x$, i.e. $\bar{p} = \argmax_{p \in P} g_{p}(x)$. We calculate the similarity map $\bar{s} = sim(\bar{p}, f(x))$ and interpolate it bilinearly to the input resolution. Then, following the visualization method from~\cite{chen2019looks}, we construct a bounding box $b(x)$ defined as the smallest rectangular region containing all activation above the $90$th percentile (see green bounding box in \cref{fig:teaser}). This region is presented to the user as the one activated by the prototypical part $\bar{p}$.

We propagate the gradient from $\bar{p}$ back to the input image using the \emph{projected gradient descent} (PGD) method~\cite{papernot2018cleverhans} to generate an adversarially modified version of the input image $\bar{x}$ (see modified image in \cref{fig:teaser}) with the goal to minimize the $g_{\bar{p}}(x)$. However, in contrast to standard PGD, we only modify the input pixels outside $b(x)$ (see adversarial modification in \cref{fig:teaser}).

If PGD manages to decrease $g_{\bar{p}}(x)$, then we deal with misalignment, as it is possible to modify activations inside $b(x)$ by modifying the region outside of it. This phenomenon is quantified using specialized \emph{spatial misalignment metrics}, which are introduced in the following paragraph.

\paragraph{Spatial misalignment metrics.}

\begin{figure}[t]
  \centering
  \includegraphics[width=.85\columnwidth]{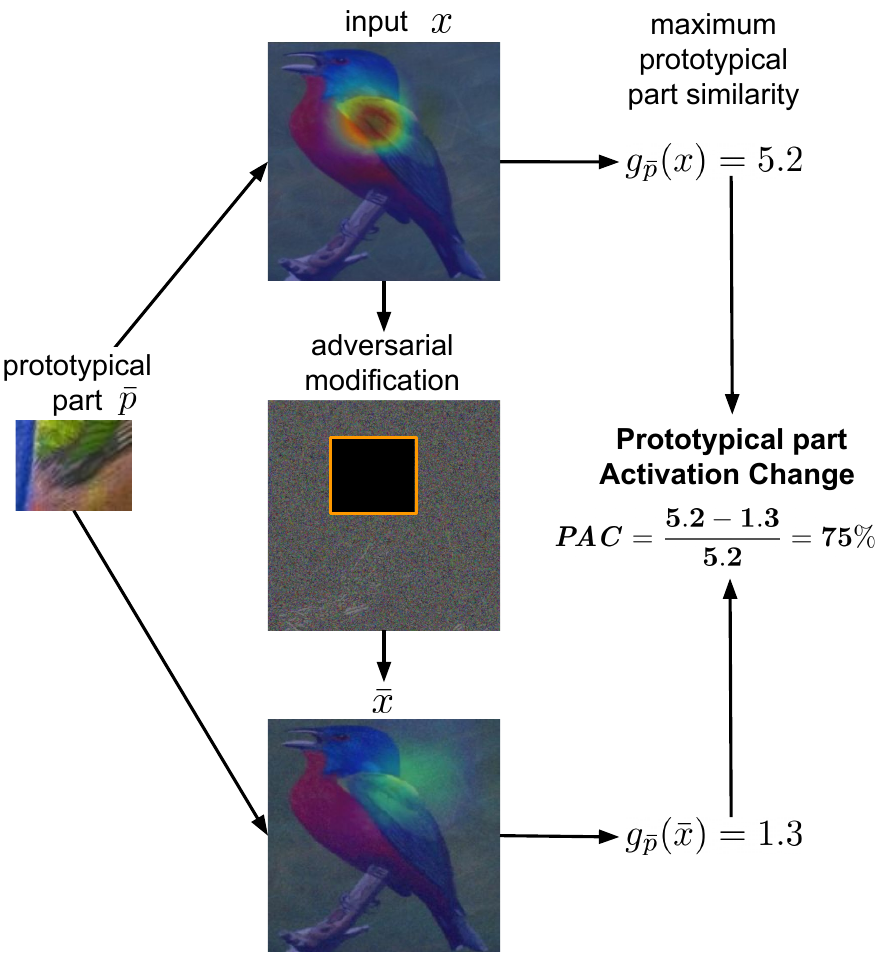}
  \caption{\textit{Prototypical part Activation Change} ($PAC$) measures the relative difference between the maximum activation of the prototypical part before and after adversarial modification. In this example, the activation of the prototypical part drops substantially (by $75\%$), which indicates a misalignment of the prototypical part explanation.}
  \label{fig:PAC}
  \vspace{-1em}
\end{figure}

\begin{figure}[t]
  \centering
  \includegraphics[width=.85\columnwidth]{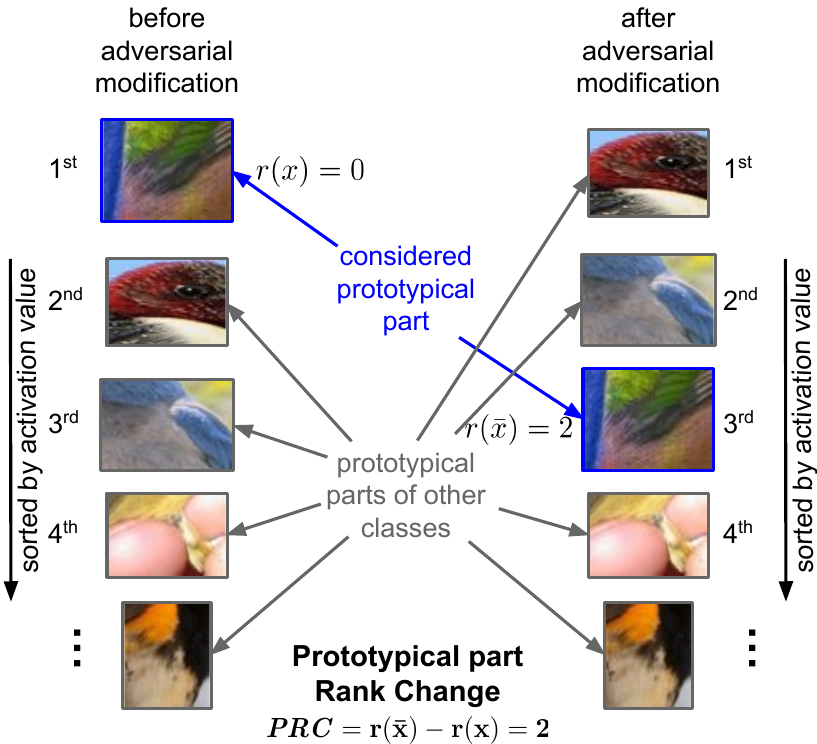}
  \caption{\textit{Prototypical part Rank Change} ($PRC$) corresponds to the difference in ranking of the prototypical part activations before and after modification. The ranking calculates the number of prototypical parts from the classes other than the class of the considered prototypical part with greater maximum similarity. $PRC$ close to $0$ indicates that the explanation is spatially aligned.}
  \label{fig:PRC}
  \vspace{-1em}
\end{figure}

In this paragraph, we describe our metrics for evaluating the spatial misalignment of prototypical parts explanations. They all operate on the original image $x$, the prototypical part $p_i$, the adversarially modified image $\bar{x_i}$, and two similarity maps ($s_i$ and $\bar{s_i}$).

The first metric, \textit{Prototypical part Location Change} ($PLC$), corresponds to the change in the explanation location (see \cref{fig:teaser})
\begin{equation}
\vspace{-0.5ex}
PLC = 1 - \mathbb{E}_{x \in X} \frac{|b(x) \cap b(\bar{x_i})|}{|b(x) \cup b(\bar{x_i})|},
\label{eq:plc}
\vspace{-0.5ex}
\end{equation}
where $b(x)$ corresponds to the minimal rectangular region covering the binarized mask obtained by assigning a positive mask value to pixels with an activation value above the $90$th percentile.
It quantifies how much the explanation region can be relocated due to changes made by adversarial modification. $PLC=0$ indicates that the region location remains unchanged. However, a high $PLC$ value suggests a significant shift in the explanation location.

The second metric, \textit{Prototypical part Activation Change ($PAC$)}, corresponds to the relative difference between the maximum activation of the prototypical part before and after adversarial modification (see~\cref{fig:PAC})
\begin{equation}
PAC = \mathbb{E}_{x \in X} \frac{g_{p_i}(x) - g_{p_i}(\bar{x_i})}{g_{p_i}(x)}.
\label{eq:pac}
\end{equation}
It quantifies the impact of adversarial modification on prototypical part activation. $PAC=0$ indicates no activation change, while higher $PAC$ values correspond to high activation changes.

The third metric, \textit{Prototypical part Rank Change}, corresponds to the difference in ranking of the prototypical parts activations (see~\cref{fig:PRC})
\begin{equation}
  PRC = \mathbb{E}_{x \in X} [r(\bar{x_i}) - r(x)],
\label{eq:prc}
\end{equation}
where $r(x) = |\{ p \in P \setminus P_k\colon g_{p}(\bar{x_i}) > g_{p_i}(\bar{x_i}) \}|$ calculates the number of prototypical parts $p$ from the classes other than the ground truth class of $x$ (here noted as class $k$) with maximum activation greater than this obtained by $\bar{p}$.
$PRC=0$ means that $\bar{p}$ remains the most activated prototypical part. However, the higher the $PRC$, the more prototypical parts from other classes become increasingly important, indirectly indicating misalignment.

As the final metric, we define the \textit{Accuracy Change} (AC), which is equal to the difference between the accuracy obtained for the original and modified images, expressed in percentage points.

\begin{figure*}[t]
  \centering
  \includegraphics[width=0.8\textwidth]{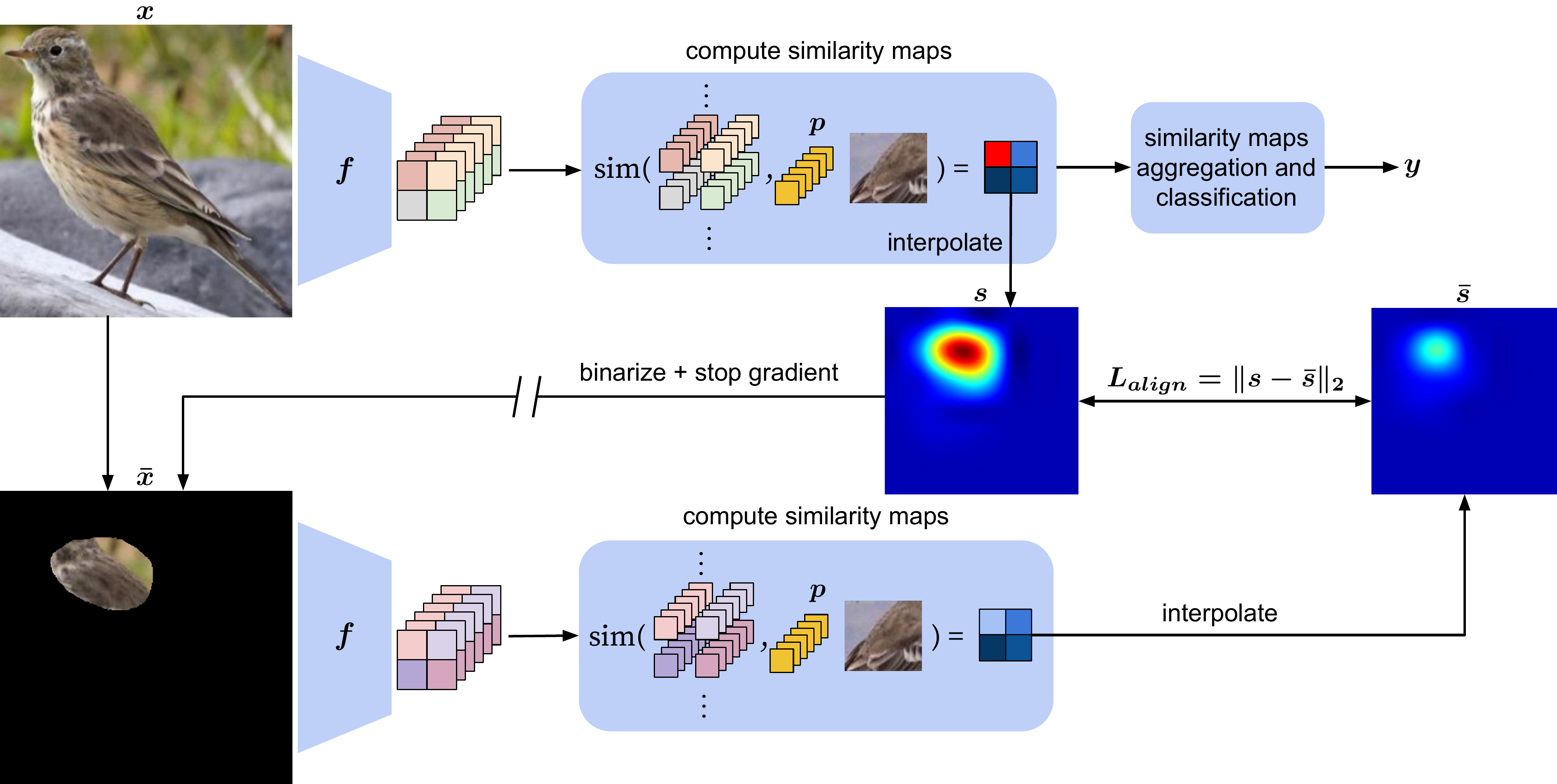}
  \caption{The spatially-aligned training aims to maximize the alignment of activation maps of a selected prototypical part between the original image and the image with only the highest activated fragment visible to the model. As shown in the picture, the training step on a single image consists of two passes of the model. During the first pass, we compute the model output together with the intermediate similarity maps to the prototypical parts. In the second pass, we randomly select a prototypical part from the ground-truth class and use its similarity map to mask the image ($\bar{x}$). This way, we obtain a new similarity map ($\bar{s}$), which we compare with the original one, obtaining loss $L_{align}$.}
  \label{fig:locality_preserving_training}
  \vspace{-1em}
\end{figure*}

\section{Misalignment compensation}

In this section, we propose a compensation methodology to prevent spatial misalignment of prototypical parts explanations. It is a general strategy as its only assumption is that the model calculates the prototypical parts similarity map over the full input image at some point in its pipeline. Hence, this strategy can be used with all state-of-the-art models based on prototypical parts.
The main idea behind our training strategy is to enforce the alignment by passing the image through the network twice: the original image and the image with the area outside the activation region masked.


To formalize our approach, let us assume that the considered prototypical part model consists of a backbone convolutional network $f$ used to calculate the similarity map $s_i = sim(p_i, f(x))$ to a prototypical part $p_i$, as defined in~\cref{eq:s}. Similar to the standard prototypical parts-based approaches, the similarity maps are aggregated and classified, as presented in Fig.~\ref{fig:locality_preserving_training}.
Additionally, we interpolate bilinearly the detached $s_i$ to the input resolution and binarize it so that positive values correspond to activation value above the 90th percentile, obtaining $m_i$. This mask is used to generate a masked input $\bar{x_i} = x \cdot m_i$ that is used in the second pass of the model. By passing $\bar{x_i}$ trough $f$ and comparing it to the prototypical part $p_i$, we obtain a new similarity map $\bar{s_i} = sim(\bar{p_i}, f(x))$.

\begin{figure}[t]
  \centering
  \includegraphics[width=1.0\columnwidth]{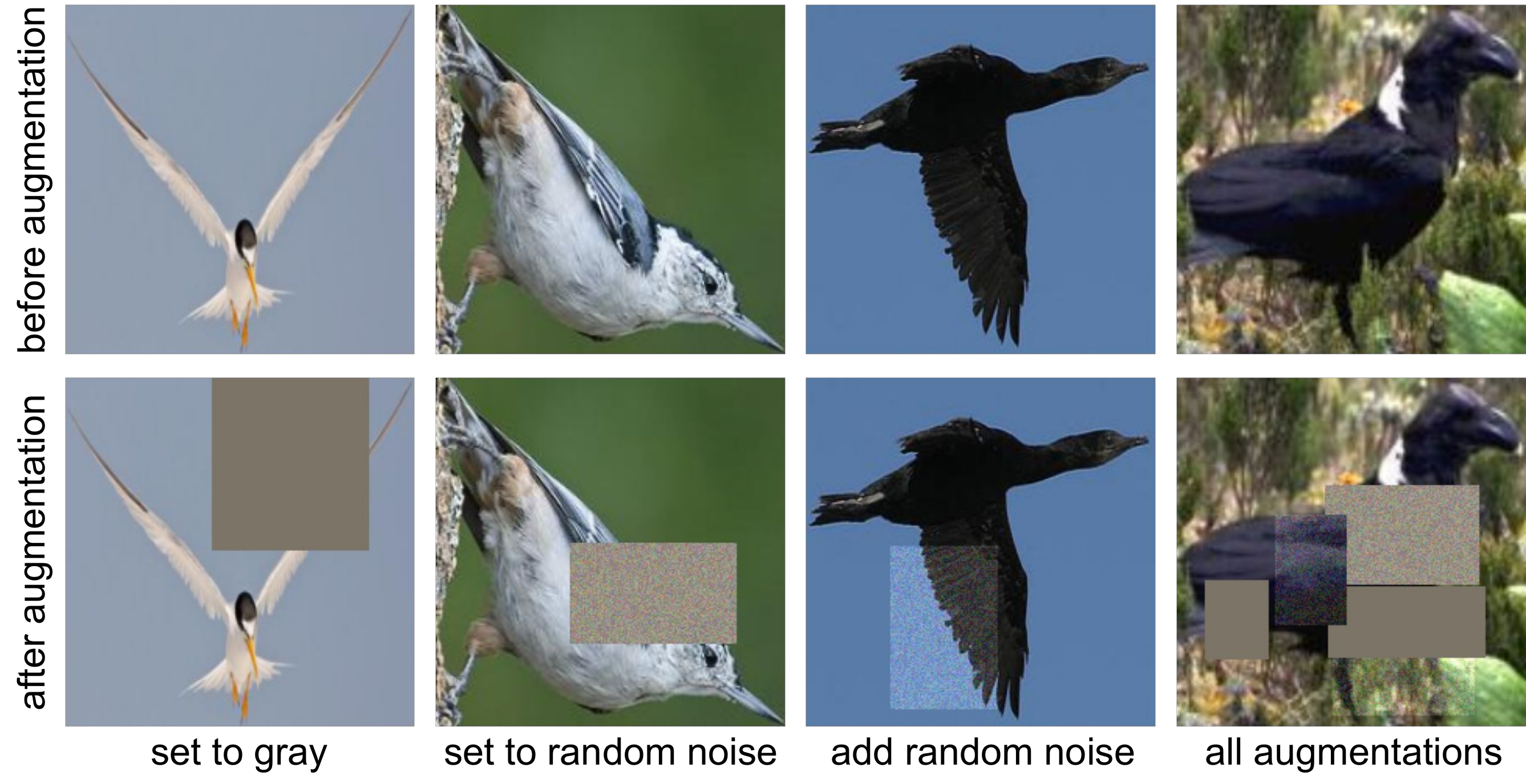}
  \caption{Sample images without (top row) and with (bottom row) the masking augmentation. Each of the left three columns presents only one type of masking, while the column on the right combines all of them. The number and types of masking augmentations are randomized. The masking augmentation steers the prototypical part model towards learning more locally-focused prototypical parts.}
  \label{fig:masking_augmentation}
  \vspace{-4ex}
\end{figure}

For aligned explanation, $\bar{s_i}$ should be very similar to $s_i$ because the area outside the activation region should not influence the final results. Therefore, we introduce a spatial alignment loss function, which penalizes the model for not fulfilling this condition
\begin{equation}
\vspace{-0.5ex}
L_{align} = \|s_i - \bar{s_i}\|_2,
\label{eq:l_align}
\end{equation}
\vspace{-0.5ex}
We weight this loss component with $\lambda_{align}$.

\begin{figure*}[t]
  \centering
  \includegraphics[width=0.9\textwidth]{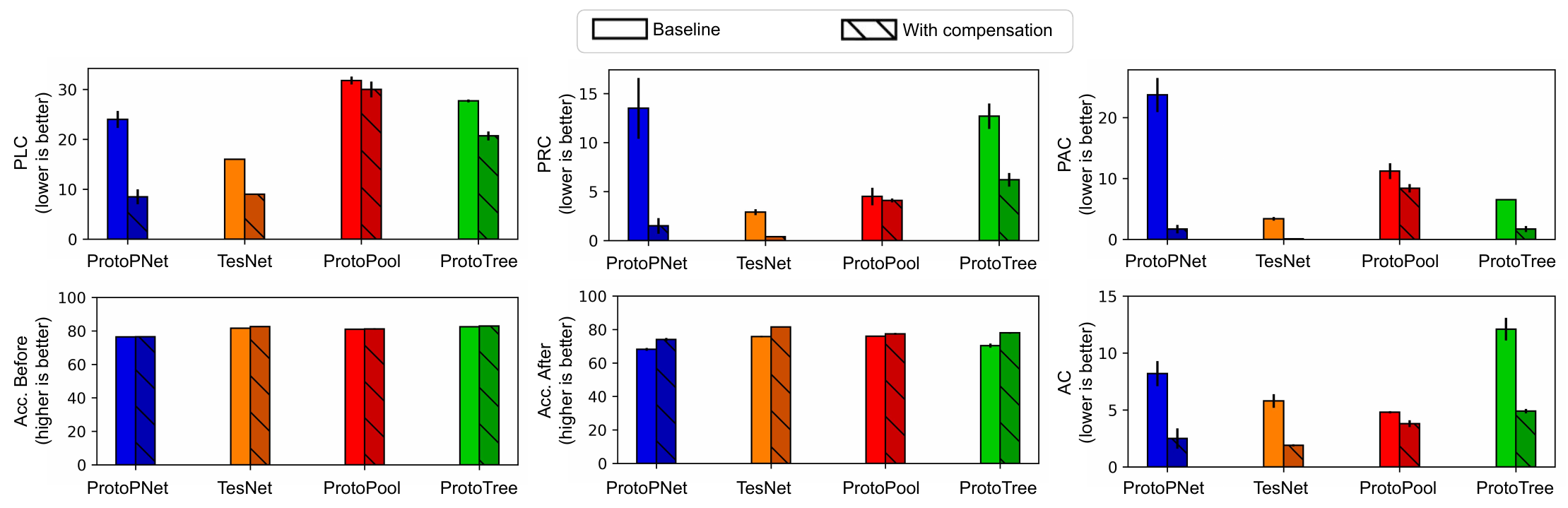}
  \caption{Comparison of the spatial misalignment metrics between baseline prototype-based models (non-hatched bars) and the best variants achieved using the spatial misalignment compensation methods (hatched bars).}
  \label{fig:main_barplot}
  \vspace{-1em}
\end{figure*}

\paragraph{Masking Augmentation.}
To further prevent explanation misalignment, we consider a special type of augmentation during model training (\cref{fig:masking_augmentation}). For every training image, we apply masking with a given probability. For the modified samples, we randomly select a number of rectangular regions together with their widths, heights, and locations. We randomly modify each region, either by adding noise or by replacing the region with gray color or random noise.

\section{Experimental setup}

In this section, we discuss the experimental setup of our spatial misalignment compensation. We describe other details related to training the compared models in Supplement. 

\paragraph{Masking augmentation setup.} For the variants of the trained models that employ masking augmentation, we apply it during all phases of the training, augmenting each sampled training image with the probability of 50\%. We sample the number of modified image regions between $1$ and $6$. For each region, we sample the augmentation type out of the three options. The width and the height of each region are randomly selected between $0.1$ and $0.5$ of the image width and height, respectively. The location of each region is randomly selected from each possible location that is fully within the image. All random values are sampled from the uniform probability on the respective intervals.

\paragraph{Spatial misalignment benchmark setup.} To evaluate the spatial misalignment of the tested models, we perform the spatial misalignment test on each image from the test set of CUB-200-2011 dataset~\cite{wah2011caltech}. For each image, we select the top activated prototypical part in the image for a given model and modify it adversarially according to the procedure described in~\cref{sec:benchmark}. We use the following parameters for the \textit{projected gradient descent} function used within the benchmark: maximum total perturbation: 0.4; maximum perturbation within one iteration: 0.01; number of iterations: 40. 

\section{Results}



\paragraph{What is the level of the spatial misalignment for vanilla models and how it can be decreased with our compensation method?}

~\cref{fig:main_barplot} illustrates the values for spatial misalignment metrics and classification accuracy (in percents) achieved by the ProtoPNet, TesNet, ProtoPool, and ProtoTree models, when trained with and without spatial-misalignment compensation on the CUB-200-2011 dataset. We show the results for baseline models and the best variants achieved with our compensation methods. The implementation of our training technique yields a notable enhancement in the robustness of the explanations, as gauged by the proposed metrics, as well as increased stability in prediction accuracy across all tested prototypical parts-based models. More details are in the Supplement. 


Relative improvements, relative to each model's baseline, exhibit varying degrees of prominence. Notably, the highest gain is observed for ProtoPNet, whereas ProtoPool demonstrates the least one. This can be related to the ProtoPool's specific focal similarity function, which is designed to generate salient explanations. Comparatively, the enhancements for ProtoTree align more closely with those observed for ProtoPNet which correlates with the findings of~\cite{nauta2023pip} emphasizing ProtoTree's limitations in capturing atomic parts of objects as prototypical parts. In the case of TesNet, its basic version presents robust interpretations, particularly evident when considering the $PAC$ metric. A comprehensive analysis of TesNet's robustness is provided in the subsequent paragraphs.

Furthermore, ProtoPNet's explanations exhibit the highest susceptibility to spatial misalignment (with the exception of the $PLC$ metric), while TesNet's interpretations show the least vulnerability to misalignment.

In terms of computations, masking augmentation has minimal impact on model training. On the other hand, computing the spatial alignment loss necessitates an additional pass, leading to an average 40\% increase in training time. Potentially, calculating the loss using only a subset of the training dataset or specific image subregions may address it.


\begin{figure}[t]
  \centering
  \includegraphics[width=0.85\columnwidth]{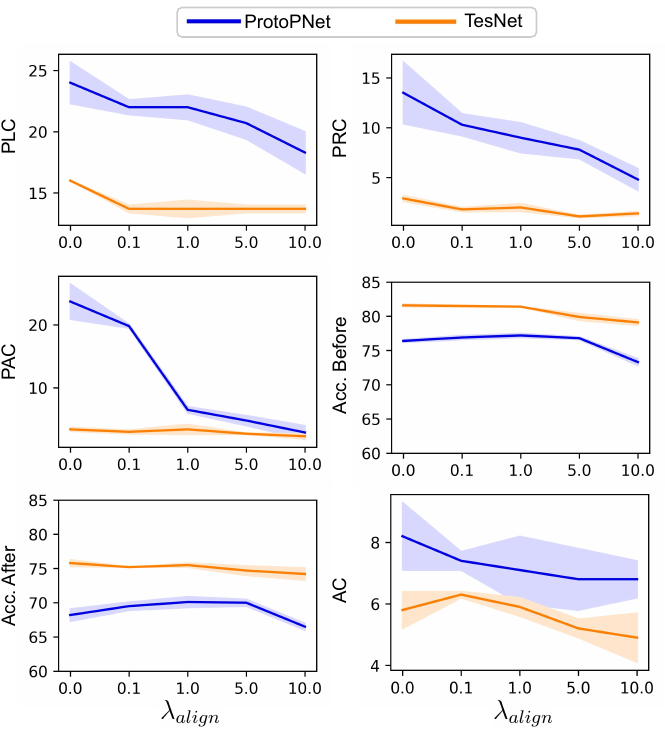}
  \caption{Ablation on the value of $\lambda_{align}$ for ProtoPNet and TesNet models, trained without the masking augmentation.}
  \label{fig:ablation_lambda}
  \vspace{-1em}
\end{figure}

\begin{figure}[t]
  \centering
  \includegraphics[width=0.9\columnwidth]{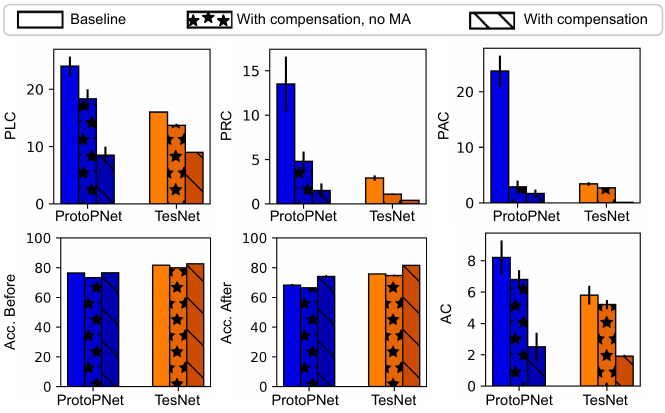}
  \caption{Ablation on the spatial-alignment compensation without and with the masking augmentation technique.}
  \label{fig:albation_masking}
  \vspace{-1em}
\end{figure}

\begin{figure}[t]
  \centering
  \includegraphics[width=0.85\columnwidth]{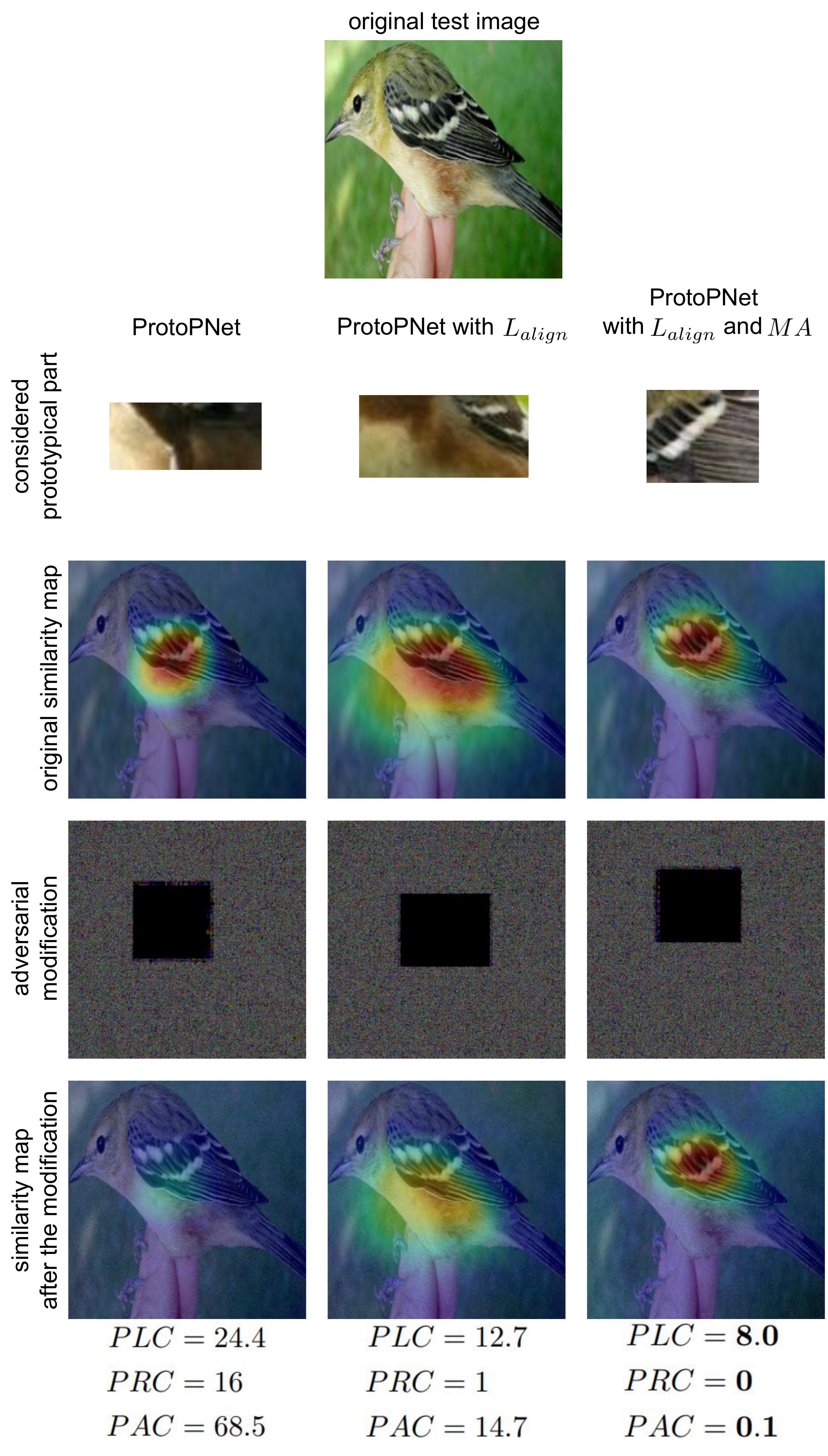}
  \caption{Comparison of the spatial misalignment benchmark results on the baseline ProtoPNet model and the variants with the spatial-aligning loss ($\lambda=10$) and with the spatial-aligning loss enhanced by the additional masking augmentation technique. Our variants achieve better consistency between the similarity map before and after modifying the image, indicating better spatial alignment.}
\label{fig:results_examples_ppnet}
\end{figure}

\paragraph{How do the explanations differ between the baselines and improved models?} In Fig.~\ref{fig:results_examples_ppnet}, we show the results of the spatial misalignment benchmark for the baseline ProtoPNet model, as well as for the ProtoPNet model trained with $\lambda_{align}=10$ with and without the masking augmentation technique. The examples were selected at random from the test set of the CUB-200-2011 dataset. We observe that the activation maps of the baseline model are diminished by the test, while the model trained with the spatial-aligning loss is robust to the modification of the image area outside the high-activation bounding box. Better spatial alignment of our improved models is also indicated by the results of the metrics shown below the images in the figure. We present more such examples in the Supplementary Materials.

\paragraph{What is the optimal weight for spatial alignment loss?}


To investigate the optimal value of the weighting factor for spatial alignment loss, we trained the models varying the value of $\lambda_{align} \in \{0.0, 0.1, 1, 5, 10\}$. For this ablation, we turned off the masking augmentation technique. The results presented in Fig.~\ref{fig:ablation_lambda} show that a larger value of the $\lambda_{align}$ weight allows obtaining more spatially-aligned explanations, as evidenced by the decreasing metric's values, with observable improvements for large weights ($\lambda_{align} \geq 1$). 
We provide more detailed results in the Supplement.

\paragraph{What is the gain from using masking augmentation?}


Fig.~\ref{fig:albation_masking} shows how training with and without masking augmentation (MA) influences the spatial misalignment metrics and the model's accuracy. We observe that, while applying the compensating loss improves the values of metrics, the additional usage of masking augmentation combined with the compensating loss yields the best results. We provide more detailed results in Supplementary Materials.


\paragraph{Does the spatial misalignment compensation generalize to other model backbones and datasets?} To evaluate the generalization of our approach, we conduct experiments using the ProtoPNet~\cite{chen2019looks} and TesNet~\cite{wang2021interpretable} models with VGG16~\cite{simonyan2014deep} backbone (instead of ResNet), as well as we benchmark the approach using the Stanford Cars dataset~\cite{krause2013stanfordcars}. With the VGG16 backbone, both models show enhanced spatial misalignment metrics, likely attributed to the narrower receptive field of VGG. Moreover, similar trends for spatial misalignment are observed on the Stanford Cars dataset, mirroring the behavior on CUB. More comprehensive results are in the Supplement.

\paragraph{Why is TesNet so robust?}

In order to investigate what makes TesNet so robust to the adversarial modifications of our benchmark, we trained it without its specific loss terms and applied the benchmark. Specifically, we trained TesNet 1) without the subspace orthogonality loss ($\lambda_{orth}$), and 2) without both the $\lambda_{orth}$ and the subspace-separation loss ($\lambda_{ss}$). Results of these experiments are provided in Supplementary Materials. Classification accuracy as well as metrics for these two models are comparable to those obtained for the baseline TesNet model. $PAC$ metric tends even to be slightly better than for the baseline TesNet model. These results might suggest that it is the prototype similarity function used by TesNet, i.e., projection of the latent space patches onto the prototype vectors (instead of a function of $L^2$-distance as used by ProtoPNet), that is primarily responsible for the superior performance of TesNet in our benchmark, as compared to results obtained by the ProtoPNet model.


\section{Conclusions}

In this article, we discuss the limitations of prototypical parts-based methods, such as ProtoPNet, caused by the misalignment between input and representation space. To address this issue, we propose an interpretability benchmark that measures this misalignment and introduce a novel compensation methodology. Experimental evaluations show the adequacy of the proposed benchmark and the effectiveness of the compensation methodology. With the proposed spatial misalignment benchmark, we can automatically assess the accuracy of explanations before presenting them to the user, thus avoiding potential misinformation.

We hope this benchmark will improve the faithfulness of visualizations generated by the prototypical parts-based models and strengthen research dedicated to the automatic assessment of models' interpretability.

\section*{Acknowledgements}
This research was partially funded by the National Science Centre, Poland, grants no.   2021/41/B/ST6/01370 (work by Mikołaj Sacha and Jacek Tabor), 2022/45/N/ST6/04147 (work by Dawid Rymarczyk), 2020/39/D/ST6/01332 (work by \L{}ukasz Struski), and 2022/47/B/ST6/03397 (work by Bartosz Zieli\'nski). The research of Bartosz Jura was carried out within the research project “Bio-inspired artificial neural network” (grant no. POIR.04.04.00-00-14DE/18-00) within the Team-Net program of the Foundation for Polish Science co-financed by the European Union under the European Regional Development Fund. Moreover, Dawid Rymarczyk received an incentive scholarship from the funds of the program Excellence Initiative -- Research University at the Jagiellonian University in Kraków.
Finally, some experiments were performed on servers purchased with funds from a Priority Research Area (Artificial Intelligence Computing Center Core Facility) grant under the Strategic Programme Excellence Initiative at Jagiellonian University.

\bigskip

\bibliography{aaai24}


\clearpage
\onecolumn
\section*{Interpretability Benchmark for Evaluating Spatial Misalignment - Supplementary Material}

\section*{Preliminaries: ProtoPNet training}
\paragraph{Training.} 
ProtoPNet learning is conducted in three optimization phases: warm-up, joint learning, and convex optimization of the last layer. In the first phase, the prototypical part layer $g$ is trained. In the second phase, $g$ and the backbone network $f$ are trained. In the last phase, the fully-connected $h$ layer is fine-tuned.
The cross-entropy loss with two regularizers, cluster and separation costs~\cite{chen2019looks}, is used as the learning criterion. Cluster cost enforces that for each training image, there is a latent patch close to at least one prototypical part of its ground truth class. On the other hand, the separation cost enforces every latent patch of an input image to be far away from the prototypical parts of other classes.

\section*{Details of the Experimental Setup}

We run our experiments on several standard protypical part-based models for image classification: ProtoPNet~\cite{chen2019looks}, TesNet~\cite{wang2021interpretable}, ProtoPool~\cite{rymarczyk2022interpretable} and ProtoTree~\cite{nauta2021neural}\footnote{We use the code provided by the authors for training the models: \\ https://github.com/cfchen-duke/ProtoPNet \\ https://github.com/JackeyWang96/TesNet \\ https://github.com/gmum/ProtoPool \\ https://github.com/M-Nauta/ProtoTree}. We train and evaluate the models on the CUB-200-2011~\cite{wah2011caltech} dataset. Apart for adding the spatial alignment compensation techniques, we use the same training procedure as in the original publications describing the models. We run multi-phase training with \textit{warmup}, \textit{joint training}, \textit{prototypical part projection}, \textit{prototypical part pruning} and \textit{last layer fine-tuning} phases~\cite{chen2019looks}, and employ \textit{resnet34}~\cite{he2016deep} network pretrained on ImageNet~\cite{deng2009imagenet} as the model backbone. We train each of the models for $\lambda_{align} \in \{0, \allowbreak 0.1, \allowbreak 1, \allowbreak 5, \allowbreak 10\}$ and with or without the masking augmentation training (MA), yielding 10 variants for each of the architectures. For each variant, we run three training runs with different random number generator seeds and take the average of their metrics as the final result. For the variants with non-zero $\lambda_{align}$, we add this loss with the weight of $\lambda_{align}$ to the loss value used by the original prototypical part-based model. 
\section*{Visual examples of spatial alignment compensation}

In this section, we show more examples that compare the spatial alignment benchmark results on the baseline ProtoPNet to the ProtoPNet with spatial aligning loss ($\lambda=10$) and with the spatial-aligning loss enhanced by the additional masking augmentation technique.


\begin{figure*}[t]
  \centering
  \includegraphics[width=\textwidth]{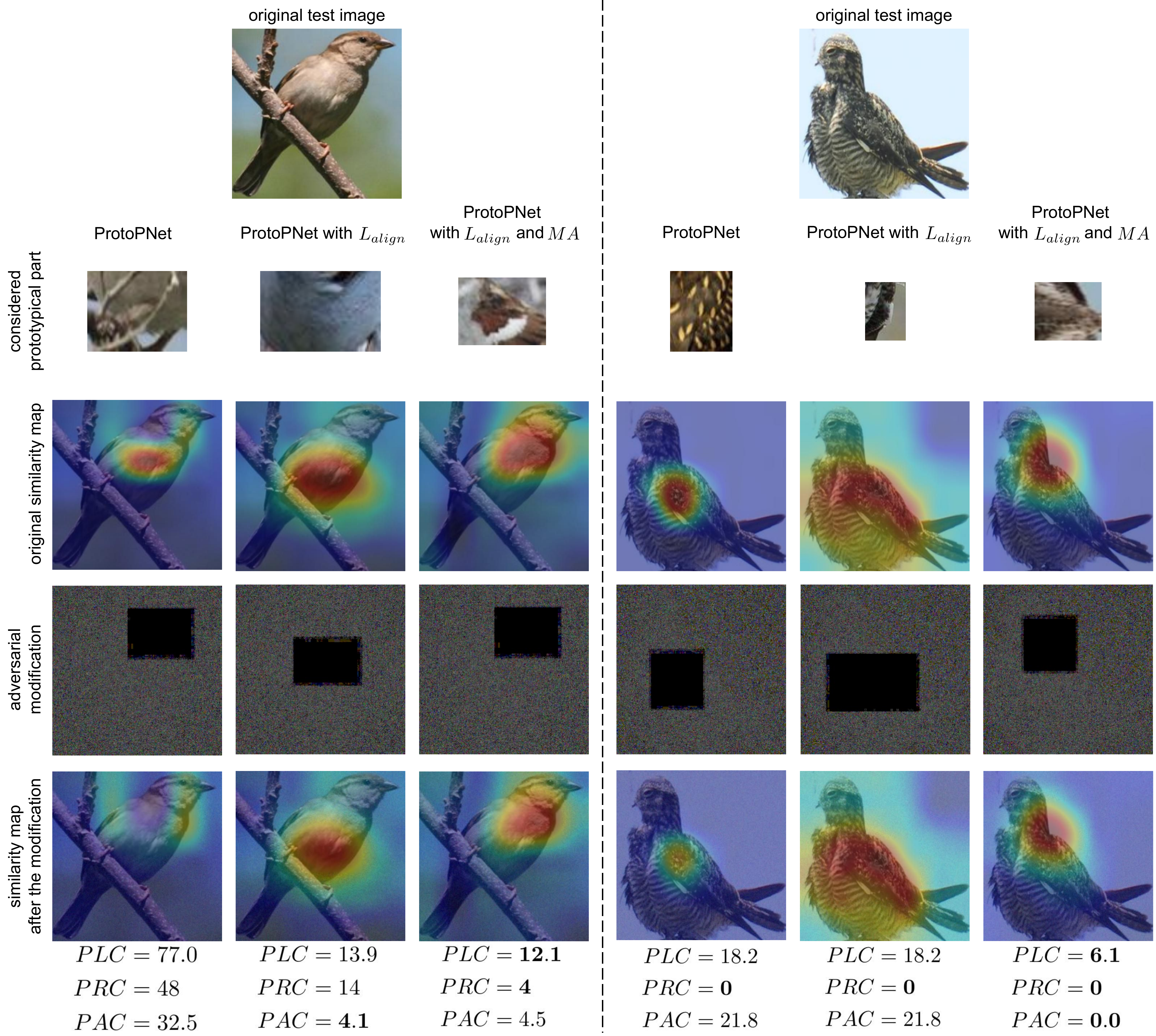}
  \label{fig:results_examples_ppnet2}
\end{figure*}

\begin{figure*}
  \centering
  \includegraphics[width=\textwidth]{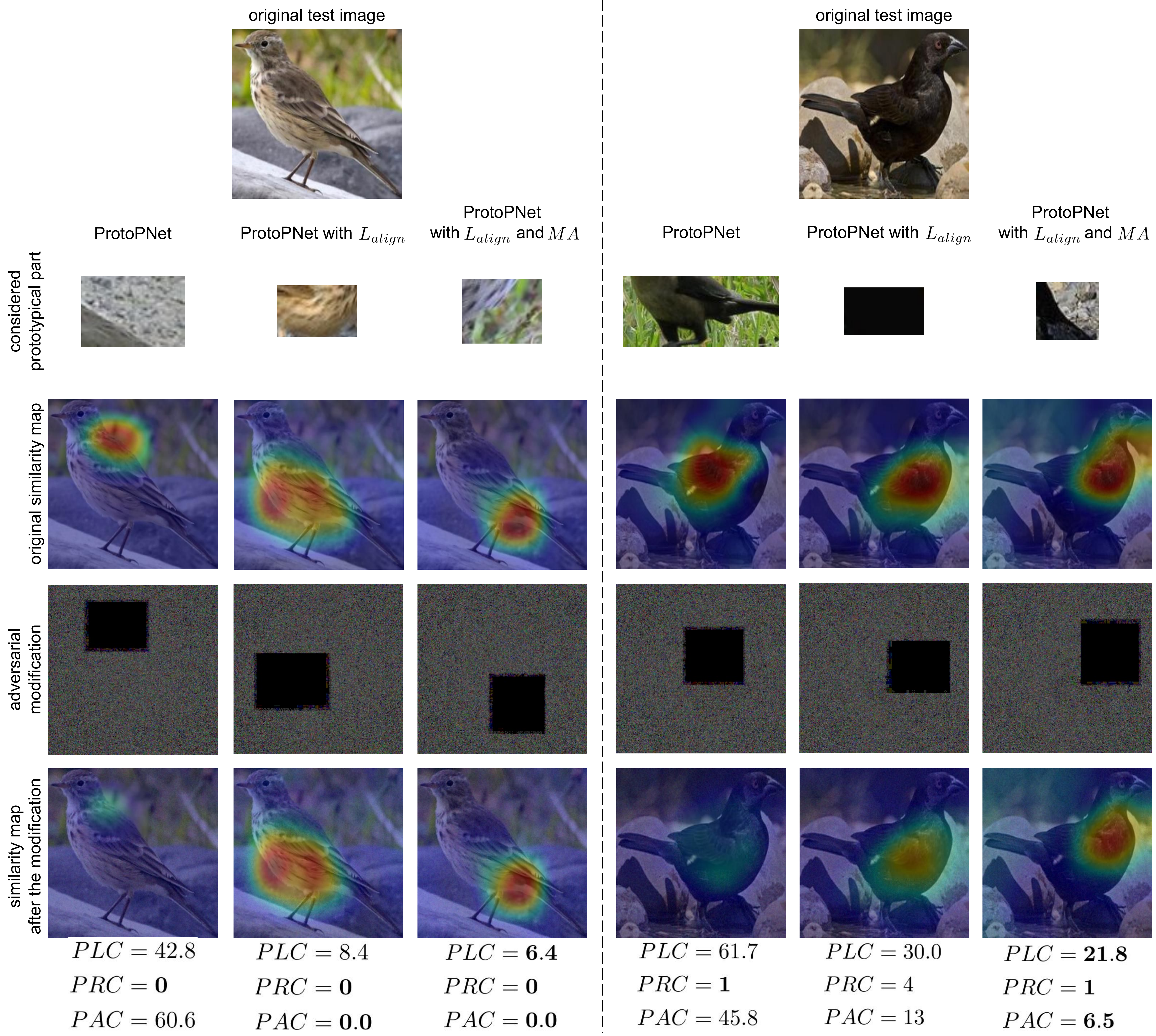}
  \label{fig:results_examples_ppnet3}
\end{figure*}

\begin{figure*}
  \centering
  \includegraphics[width=\textwidth]{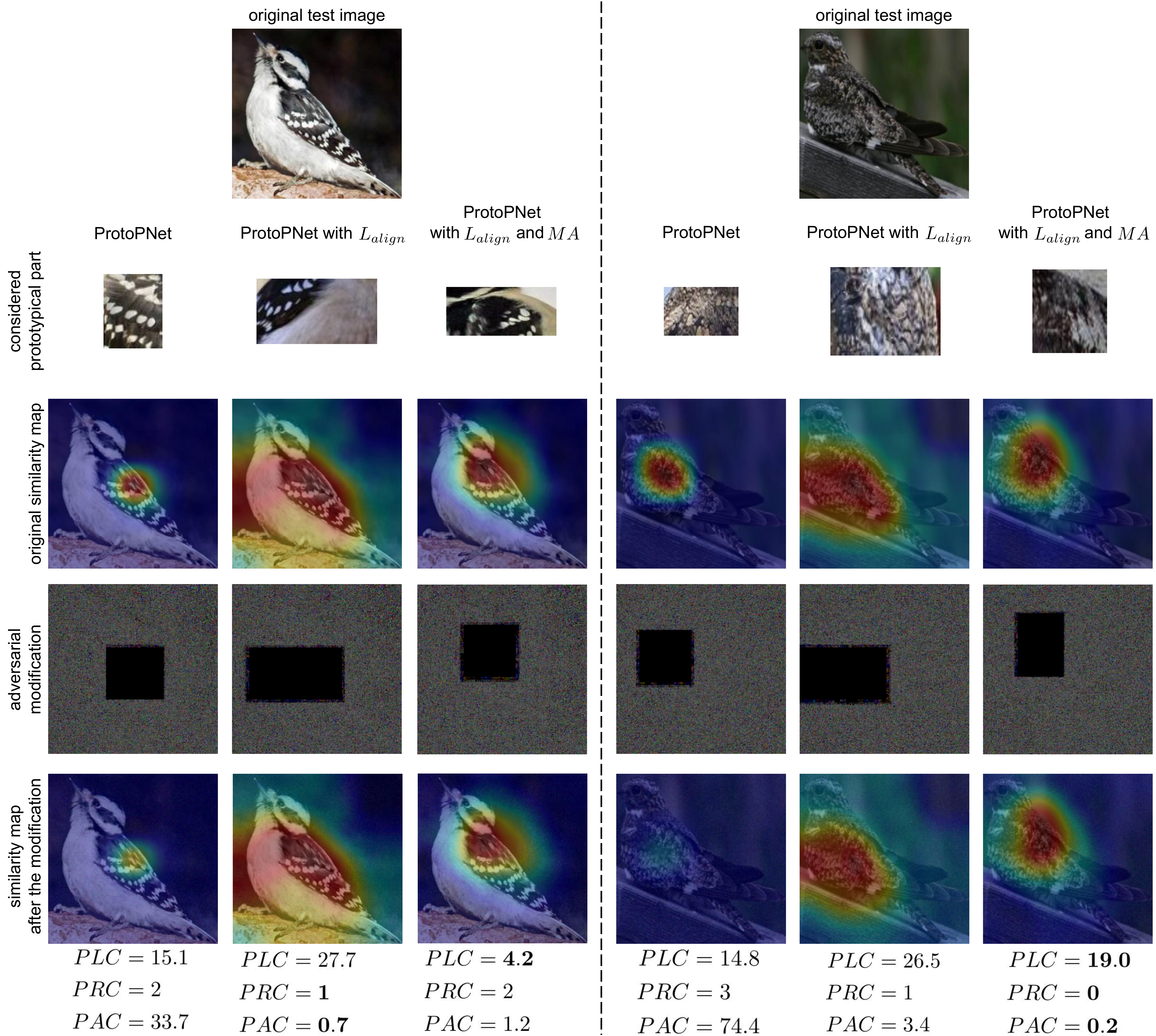}
  \label{fig:results_examples_ppnet4}
\end{figure*}


\onecolumn
\section*{Detailed results of spatial-aligning compensation}
In the following sections, we show the detailed comparison of spatial alignment metrics between baseline methods and methods improved by the spatial-aligning loss and masking augmentation. For each of the evaluated models, we show the values of Prototypical part Location Change (PLC), Prototypical part Rank Change (PRC), Prototypical part Activation Change (PAC), as well as the classification accuracy before (Acc. Before), and after (Acc. After) the adversarial modification and the difference in percentage points between these accuracy values (Accuracy Change - AC).

\subsection{Statistics over the values of metrics}
\begin{figure}[t]
  \centering
  \includegraphics[width=0.8\columnwidth]{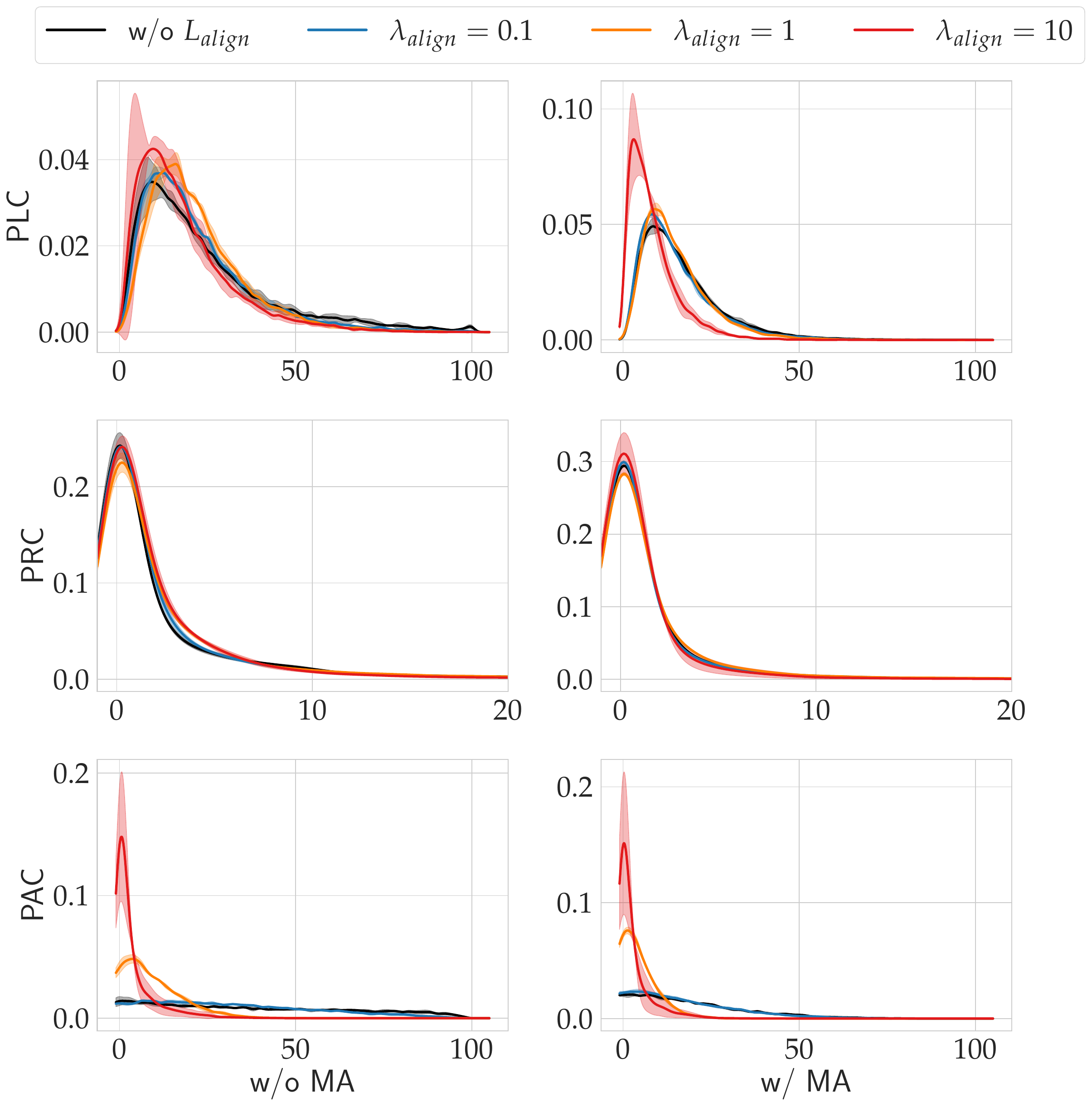}
  \caption{Kernel density estimation over histograms of metrics values based on the test images from CUB-200-2011 for the ProtoPNet variants trained with different values of $\lambda_{align}$ as well as with and without the masking augmentation (MA). We observe that, on average, our variants achieve better (lower) metrics values, getting better with both increasing $\lambda_{align}$ and MA.}
  \label{fig:results_kde}
\end{figure}

In~\cref{fig:results_kde}, we draw kernel density estimation plots over histograms of all values of the spatial alignment metrics achieved by the trained ProtoPNet models on the test set of CUB-200-2011. We observe that, on average, our variants achieve better (lower) values of metrics, with the values of the metrics getting better with both increasing $\lambda_{align}$ and employing the masking augmentation.

\clearpage

\subsection{Detailed results on different prototypical part-based models}
In \cref{tab:results_ppnet,tab:results_tesnet,tab:results_protopool,tab:results_prototree} (\cite{chen2019looks,wang2021interpretable,rymarczyk2022interpretable,nauta2021neural}), we compare the values of spatial alignment metrics between baseline methods and methods improved by the spatial-aligning loss and masking augmentation.

\vskip 0.2in

\begin{table*}[t]
    \centering
    \resizebox{\textwidth}{!}{%
    \begin{tabular}{lrc|rrr|ccc}
         \toprule
         method & $\lambda_{align}$ & \multicolumn{1}{c}{MA} & \multicolumn{1}{c}{$\downarrow$ PLC } & \multicolumn{1}{c}{$\downarrow$ PRC} & \multicolumn{1}{c}{$\downarrow$ PAC } & $\uparrow$ Acc. Before & $\uparrow$ Acc. After & $\downarrow$ AC \\
         \toprule
         baseline & \xmark & \xmark & 24.0 $\pm$ 1.7 & 13.5 $\pm$ 3.1 & 23.7 $\pm$ 2.8 & 76.4 $\pm$ 0.2 & 68.2 $\pm$ 0.9 & 8.2 $\pm$ 1.1 \\
         \midrule
         \multirow{7}{*}{ours} & 0.1 & \multirow{3}{*}{\xmark} & 22.0 $\pm$ 0.6 & 10.3 $\pm$ 1.1 & 19.8 $\pm$ 0.3 & 76.9 $\pm$ 0.3 & 69.5 $\pm$ 0.6 & 7.4 $\pm$ 0.3 \\
         & 1  & & 22.0 $\pm$ 1.0 & 9.0 $\pm$ 1.5 & 6.5 $\pm$ 0.5 & 77.2 $\pm$ 0.3 & 70.1 $\pm$ 0.8 & 7.1 $\pm$ 1.1 \\
         & 5  & & 20.7 $\pm$ 1.3 & 7.8 $\pm$ 0.9 & 4.8 $\pm$ 0.8 & 76.8 $\pm$ 0.2 & 70.0 $\pm$ 0.5 & 6.8 $\pm$ 1.0 \\
         & 10 & & 18.3 $\pm$ 1.7 & 4.8 $\pm$ 1.1 & 2.9 $\pm$ 1.1 & 73.3 $\pm$ 0.5 & 66.5 $\pm$ 0.5 & 6.8 $\pm$ 0.6 \\
         \cmidrule{2-9}
         & 0 & \multirow{4}{*}{\cmark} & 16.7 $\pm$ 0.3 & 1.8 $\pm$ 0.1 & 5.0 $\pm$ 0.7 & 78.6 $\pm$ 0.6 & 77.0 $\pm$ 0.7 & 1.6 $\pm$ 0.3 \\
         & 0.1 & & 15.7 $\pm$ 0.3 & 1.7 $\pm$ 0.1 & 6.3 $\pm$ 0.2 & 78.1 $\pm$ 0.2 & 75.9 $\pm$ 0.2 & 2.2 $\pm$ 0.1 \\
         & 1 & & 15.3 $\pm$ 0.3 & 2.1 $\pm$ 0.1 & 2.7 $\pm$ 0.1 & 76.7 $\pm$ 0.2 & 74.7 $\pm$ 0.1 & 2.0 $\pm$ 0.2 \\
         & 5 & & 13.3 $\pm$ 0.8 & 1.8 $\pm$ 0.4 & 2.5 $\pm$ 0.3 & 76.7 $\pm$ 0.2 & 74.6 $\pm$ 0.3 & 2.1 $\pm$ 0.4 \\
         & 10 & & 8.5 $\pm$ 1.5 & 1.6 $\pm$ 0.8 & 1.7 $\pm$ 0.7 & 76.5 $\pm$ 0.1 & 74.0 $\pm$ 1.0 & 2.5 $\pm$ 0.9 \\
         \bottomrule
    \end{tabular}}
    \caption{Metrics on the spatial-alignment test acquired by the baseline ProtoPNet model and ProtoPNet trained with the spatial-aligning training and masking augmentation. The variant with the spatially-aligning loss with weight $\lambda_{align}=10$ and masking augmentation achieves overall the best results, and it was used in the main paper results.}
    \label{tab:results_ppnet}
\end{table*}

\begin{table*}[t]
    \centering
    \resizebox{\textwidth}{!}{%
    \begin{tabular}{lrc|rrr|ccc}
         \toprule
         method & $\lambda_{align}$ & \multicolumn{1}{c}{MA} & \multicolumn{1}{c}{$\downarrow$ PLC } & \multicolumn{1}{c}{$\downarrow$ PRC} & \multicolumn{1}{c}{$\downarrow$ PAC } & $\uparrow$ Acc. Before & $\uparrow$ Acc. After & $\downarrow$ AC \\
         \toprule
         baseline & \xmark & \xmark & 16.0 $\pm$ 0.0 & 2.9 $\pm$ 0.3 & 3.4 $\pm$ 0.3 & 81.6 $\pm$ 0.2 & 75.8 $\pm$ 0.5 & 5.8 $\pm$ 0.6 \\
         \midrule
         \multirow{9}{*}{ours} & 0.1 & \multirow{4}{*}{\xmark} & 13.7 $\pm$ 0.3 & 1.8 $\pm$ 0.2 & 3.0 $\pm$ 0.3 & 81.5 $\pm$ 0.1 & 75.2 $\pm$ 0.0 & 6.3 $\pm$ 0.1 \\
         & 1 & & 13.7 $\pm$ 0.7 & 2.0 $\pm$ 0.4 & 3.4 $\pm$ 0.8 & 81.4 $\pm$ 0.0 & 75.5 $\pm$ 0.3 & 5.9 $\pm$ 0.3 \\
         & 5 & & 13.7 $\pm$ 0.3 & 1.1 $\pm$ 0.1 & 2.7 $\pm$ 0.1 & 79.9 $\pm$ 0.5 & 74.7 $\pm$ 0.7 & 5.2 $\pm$ 0.3 \\
         & 10 & & 13.7 $\pm$ 0.3 & 1.4 $\pm$ 0.2 & 2.3 $\pm$ 0.4 & 79.1 $\pm$ 0.4 & 74.2 $\pm$ 0.9 & 4.9 $\pm$ 0.8 \\
         \cmidrule{2-9}
         & 0 & \multirow{5}{*}{\cmark} & 9.0 $\pm$ 0.0 & 0.4 $\pm$ 0.0 & 0.1 $\pm$ 0.0 & 82.6 $\pm$ 0.2 & 81.5 $\pm$ 0.2 & 1.9 $\pm$ 0.1 \\
         & 0.1 & & 9.0 $\pm$ 0.0 & 0.4 $\pm$ 0.0 & 0.3 $\pm$ 0.1 & 82.4 $\pm$ 0.2 & 80.4 $\pm$ 0.2 & 2.0 $\pm$ 0.1 \\
         & 1 & & 9.0 $\pm$ 0.0 & 0.4 $\pm$ 0.1 & 0.8 $\pm$ 0.1 & 82.2 $\pm$ 0.2 & 80.3 $\pm$ 0.2 & 1.9 $\pm$ 0.1 \\
         & 5 & & 9.0 $\pm$ 0.0 & 0.3 $\pm$ 0.0 & 0.8 $\pm$ 0.2 & 80.7 $\pm$ 0.5 & 79.4 $\pm$ 0.6 & 1.4 $\pm$ 0.2 \\
         & 10 & & 10.0 $\pm$ 0.0 & 0.4 $\pm$ 0.0 & 0.8 $\pm$ 0.1 & 79.3 $\pm$ 0.5 & 77.1 $\pm$ 0.7 & 2.2 $\pm$ 0.3 \\
         \bottomrule
    \end{tabular}}
    \caption{Metrics on the spatial-alignment test acquired by the baseline TesNet model and TesNet trained with the spatial-aligning training and masking augmentation. The variant with no spatially-aligning loss ($\lambda_{align}=0$) and masking augmentation achieves overall the best results, and it was used in the main paper results.}
    \label{tab:results_tesnet}
\end{table*}

\begin{table*}[t]
    \centering
    \resizebox{\textwidth}{!}{%
    \begin{tabular}{lrc|rrr|ccc}
         \toprule
         method & $\lambda_{align}$ & \multicolumn{1}{c}{MA} & \multicolumn{1}{c}{$\downarrow$ PLC } & \multicolumn{1}{c}{$\downarrow$ PRC} & \multicolumn{1}{c}{$\downarrow$ PAC } & $\uparrow$ Acc. Before & $\uparrow$ Acc. After & $\downarrow$ AC \\
         \toprule
         baseline & \xmark & \xmark & 31.8 $\pm$ 0.8 & 4.5 $\pm$ 0.9 & 11.2 $\pm$ 1.3 & 80.8 $\pm$ 0.2 & 76.0 $\pm$ 0.3 & 4.8 $\pm$ 0.1 \\
         \midrule
         \multirow{7}{*}{ours} & 0.1 & \multirow{3}{*}{\xmark} & 30.5 $\pm$ 0.9 & 4.6 $\pm$ 0.3 & 11.5 $\pm$ 0.5 & 81.1 $\pm$ 0.4 & 75.9 $\pm$ 0.7 & 6.2 $\pm$ 0.3 \\
         & 1  & & 30.8 $\pm$ 1.3 & 3.7 $\pm$ 0.4 & 8.6 $\pm$ 0.6 & 81.5 $\pm$ 0.1 & 76.0 $\pm$ 0.4 & 5.5 $\pm$ 0.3 \\
         & 5  & & 31.4 $\pm$ 1.5 & 3.7 $\pm$ 0.4 & 8.7 $\pm$ 0.5 & 81.3 $\pm$ 0.1 & 76.4 $\pm$ 0.3 & 4.9 $\pm$ 0.3 \\
         & 10 & & 32.4 $\pm$ 1.5 & 3.8 $\pm$ 0.5 & 8.9 $\pm$ 0.8 & 81.3 $\pm$ 0.1 & 76.9 $\pm$ 0.5 & 4.4 $\pm$ 0.4 \\
         \cmidrule{2-9}
         & 0 & \multirow{4}{*}{\cmark} & 32.4 $\pm$ 1.0 & 3.4 $\pm$ 0.6 & 9.3 $\pm$ 1.1 & 80.9 $\pm$ 0.2 & 76.3 $\pm$ 0.2 & 4.6 $\pm$ 0.2 \\
         & 0.1 & & 31.1 $\pm$ 1.6 & 3.7 $\pm$ 0.4 & 9.8 $\pm$ 0.5 & 80.4 $\pm$ 0.4 & 76.4 $\pm$ 0.2 & 4.0 $\pm$ 0.5 \\
         & 1 & & 32.1 $\pm$ 1.9 & 3.5 $\pm$ 0.3 & 9.0 $\pm$ 0.9 & 81.5 $\pm$ 0.1 & 76.8 $\pm$ 0.3 & 4.7 $\pm$ 0.4 \\
         & 5 & & 31.5 $\pm$ 1.2 & 3.6 $\pm$ 0.4 & 8.8 $\pm$ 0.8 & 81.4 $\pm$ 0.1 & 76.4 $\pm$ 0.5 & 5.1 $\pm$ 0.5 \\
         & 10 & & 30.0 $\pm$ 1.6 & 4.1 $\pm$ 0.2 & 8.4 $\pm$ 0.7 & 81.2 $\pm$ 0.5 & 77.4 $\pm$ 0.6 & 3.8 $\pm$ 0.3 \\
         \bottomrule
    \end{tabular}}
    \caption{Metrics on the spatial-alignment test acquired by the baseline ProtoPool model and ProtoPool trained with the spatial-aligning training and masking augmentation. The variant with the spatially-aligning loss with weight $\lambda_{align}=10$ and masking augmentation achieves overall the best results, and it was used in the main paper results.}
    \label{tab:results_protopool}
\end{table*}

\begin{table*}[t]
    \centering
    \resizebox{\textwidth}{!}{%
    \begin{tabular}{lrc|rrr|ccc}
         \toprule
         method & $\lambda_{align}$ & \multicolumn{1}{c}{MA} & \multicolumn{1}{c}{$\downarrow$ PLC } & \multicolumn{1}{c}{$\downarrow$ PRC} & \multicolumn{1}{c}{$\downarrow$ PAC } & $\uparrow$ Acc. Before & $\uparrow$ Acc. After & $\downarrow$ AC \\
         \toprule
         baseline & \xmark & \xmark & 27.7 $\pm$ 0.3 & 12.7 $\pm$ 1.3 & 6.5 $\pm$ 0.1 & 82.5 $\pm$ 0.2 & 70.4 $\pm$ 1.2 & 12.1 $\pm$ 1.0 \\
         \midrule
         \multirow{9}{*}{ours} & 0.1 & \multirow{4}{*}{\xmark} & 28.0 $\pm$ 0.6 & 12.8 $\pm$ 0.6 & 6.2 $\pm$ 0.2  & 79.6 $\pm$ 2.3 & 66.5 $\pm$ 1.3 & 13.1 $\pm$ 1.4 \\
         & 1 & & 28.0 $\pm$ 0.6 & 12.3 $\pm$ 0.9 & 5.3 $\pm$ 0.5 & 79.6 $\pm$ 2.3 & 66.5 $\pm$ 1.3 & 13.1 $\pm$ 1.4 \\
         & 5 & & 24.3 $\pm$ 1.3 & 11.2 $\pm$ 0.3 & 4.7 $\pm$ 1.4 & 81.6 $\pm$ 0.2 & 66.4 $\pm$ 2.0 & 15.2 $\pm$ 2.2 \\
         & 10 & & 21.7 $\pm$ 0.9 & 12.2 $\pm$ 0.8 & 5.6 $\pm$ 1.1 & 81.2 $\pm$ 1.6 & 67.4 $\pm$ 2.8 & 13.8 $\pm$ 1.3 \\
         \cmidrule{2-9}
         & 0 & \multirow{5}{*}{\cmark} & 21.7 $\pm$ 0.3 & 8.4 $\pm$ 0.5 & 2.5 $\pm$ 0.4 & 82.9 $\pm$ 0.5 & 78.0 $\pm$ 0.6 & 4.9 $\pm$ 0.2 \\
         & 0.1 & & 20.3 $\pm$ 0.3 & 5.4 $\pm$ 0.1 & 2.9 $\pm$ 0.1 & 81.9 $\pm$ 0.2 & 77.3 $\pm$ 0.3 & 4.6 $\pm$ 0.1 \\
         & 1 & & 20.7 $\pm$ 0.9 & 6.2 $\pm$ 0.7 & 1.7 $\pm$ 0.5 & 82.9 $\pm$ 0.3 & 78.0 $\pm$ 0.5 & 4.9 $\pm$ 0.2 \\
         & 5 & & 17.7 $\pm$ 1.5 & 5.9 $\pm$ 0.5 & 2.3 $\pm$ 0.8 & 81.7 $\pm$ 0.6 & 76.6 $\pm$ 0.6 & 5.1 $\pm$ 0.1 \\
         & 10 & & 16.7 $\pm$ 0.3 & 5.7 $\pm$ 0.0 & 2.7 $\pm$ 0.3 & 72.4 $\pm$ 1.5 & 66.8 $\pm$ 1.5 & 5.7 $\pm$ 0.6 \\
         \bottomrule
    \end{tabular}}
    \caption{Metrics on the spatial-alignment test acquired by the baseline ProtoTree model and ProtoTree trained with the spatial-aligning training and masking augmentation. The variant with the spatially-aligning loss with weight $\lambda_{align}=1$ and masking augmentation achieves overall the best results, and it was used in the main paper results.}
    \label{tab:results_prototree}
\end{table*}

\clearpage

\subsection{Detailed results on VGG backbone}
In \cref{tab:results_ppnet_vgg,tab:results_tesnet_vgg}, we show the detailed comparison of spatial alignment metrics between baseline methods and methods improved by the spatial-aligning loss and masking augmentation for ProtoPNet~\cite{chen2019looks} and TesNet~\cite{wang2021interpretable} trained using VGG16 backbone instead of Resnet34. 

\vskip 0.5in

\begin{table*}[t]
    \centering
    \resizebox{\textwidth}{!}{%
    \begin{tabular}{lrc|rrr|ccc}
         \toprule
         method & $\lambda_{align}$ & \multicolumn{1}{c}{MA} & \multicolumn{1}{c}{$\downarrow$ PLC } & \multicolumn{1}{c}{$\downarrow$ PRC} & \multicolumn{1}{c}{$\downarrow$ PAC } & $\uparrow$ Acc. Before & $\uparrow$ Acc. After & $\downarrow$ AC \\
         \toprule
         baseline & \xmark & \xmark & 17.9 $\pm$ 0.4 & 1.2 $\pm$ 0.1 & 7.0 $\pm$ 0.3 & 71.0 $\pm$ 0.2 & 64.9 $\pm$ 0.3 & 4.8 $\pm$ 0.1 \\
         \midrule
         \multirow{7}{*}{ours} & 0.1 & \multirow{3}{*}{\xmark} & 16.7 $\pm$ 0.3 & 0.9 $\pm$ 0.0 & 5.4 $\pm$ 0.2 & 71.3 $\pm$ 0.1 & 65.2 $\pm$ 0.1 & 6.1 $\pm$ 0.1 \\
         & 1  & & 17.0 $\pm$ 0.5 & 1.1 $\pm$ 0.1 & 4.9 $\pm$ 0.3 & 72.7 $\pm$ 0.2 & 66.7 $\pm$ 0.5 & 5.9 $\pm$ 0.6 \\
         & 5  & & 18.2 $\pm$ 0.3 & 1.8 $\pm$ 0.1 & 5.5 $\pm$ 0.7 & 73.6 $\pm$ 0.1 & 67.8 $\pm$ 0.1 & 5.8 $\pm$ 0.1 \\
         & 10 & & 17.9 $\pm$ 0.3 & 1.8 $\pm$ 0.2 & 4.9 $\pm$ 0.7 & 73.6 $\pm$ 0.2 & 67.9 $\pm$ 0.3 & 5.7 $\pm$ 0.4 \\
         \cmidrule{2-9}
         & 0 & \multirow{4}{*}{\cmark} & 12.9 $\pm$ 0.3 & 0.5 $\pm$ 0.0 & 3.6 $\pm$ 0.1 & 73.2 $\pm$ 0.1 & 70.4 $\pm$ 0.4 & 2.8 $\pm$ 0.5 \\
         & 0.1 & & 11.8 $\pm$ 0.3 & 0.4 $\pm$ 0.0 & 2.9 $\pm$ 0.1 & 74.2 $\pm$ 0.2 & 71.4 $\pm$ 0.1 & 2.8 $\pm$ 0.2 \\
         & 1 & & 12.0 $\pm$ 0.2 & 0.5 $\pm$ 0.0 & 2.2 $\pm$ 0.1 & 74.6 $\pm$ 0.1 & 72.0 $\pm$ 0.1 & 2.5 $\pm$ 0.1 \\
         & 5  & & 12.6 $\pm$ 0.2 & 0.6 $\pm$ 0.0 & 2.6 $\pm$ 0.1 & 75.1 $\pm$ 0.3 & 72.6 $\pm$ 0.2 & 2.5 $\pm$ 0.1 \\
         & 10 & & 13.2 $\pm$ 0.2 & 0.8 $\pm$ 0.0 & 3.0 $\pm$ 0.1 & 74.8 $\pm$ 0.3 & 72.3 $\pm$ 0.2 & 2.6 $\pm$ 0.2 \\
         \bottomrule
    \end{tabular}}
    \caption{Metrics on the spatial-alignment test acquired by the baseline ProtoPNet model and ProtoPNet trained with the spatial-aligning training and masking augmentation, using VGG16 backbone instead of Resnet34.}
    \label{tab:results_ppnet_vgg}
\end{table*}

\begin{table*}[t]
    \centering
    \resizebox{\textwidth}{!}{%
    \begin{tabular}{lrc|rrr|ccc}
         \toprule
         method & $\lambda_{align}$ & \multicolumn{1}{c}{MA} & \multicolumn{1}{c}{$\downarrow$ PLC } & \multicolumn{1}{c}{$\downarrow$ PRC} & \multicolumn{1}{c}{$\downarrow$ PAC } & $\uparrow$ Acc. Before & $\uparrow$ Acc. After & $\downarrow$ AC \\
         \toprule
         baseline & \xmark & \xmark & 15.0 $\pm$ 0.6 & 0.9 $\pm$ 0.1 & 1.0 $\pm$ 0.1 & 79.3 $\pm$ 0.1 & 72.6 $\pm$ 0.2 & 6.7 $\pm$ 0.4 \\
         \midrule
         \multirow{9}{*}{ours} & 0.1 & \multirow{4}{*}{\xmark} & 15.0 $\pm$ 0.6 & 0.8 $\pm$ 0.1 & 1.2 $\pm$ 0.1 & 79.4 $\pm$ 0.2 & 72.1 $\pm$ 0.3 & 7.4 $\pm$ 0.5 \\
         & 1 & & 12.7 $\pm$ 0.3 & 0.6 $\pm$ 0.0 & 1.3 $\pm$ 0.1 & 80.0 $\pm$ 0.2 & 73.4 $\pm$ 0.4 & 6.6 $\pm$ 0.4 \\
         & 5 & & 13.3 $\pm$ 0.3 & 0.5 $\pm$ 0.1 & 1.0 $\pm$ 0.1 & 79.5 $\pm$ 0.2 & 71.9 $\pm$ 0.5 & 7.6 $\pm$ 0.5 \\
         & 10 & & 12.3 $\pm$ 0.3 & 0.3 $\pm$ 0.0 & 0.6 $\pm$ 0.0 & 78.5 $\pm$ 0.1 & 71.7 $\pm$ 0.4 & 6.8 $\pm$ 0.4 \\
         \cmidrule{2-9}
         & 0 & \multirow{5}{*}{\cmark} & 9.0 $\pm$ 0.0 & 0.4 $\pm$ 0.0 & 0.4 $\pm$ 0.1 & 80.4 $\pm$ 0.1 & 78.0 $\pm$ 0.3 & 2.3 $\pm$ 0.2 \\
         & 0.1 & & 9.0 $\pm$ 0.0 & 0.3 $\pm$ 0.0 & 0.4 $\pm$ 0.1 & 80.5 $\pm$ 0.2 & 78.3 $\pm$ 0.2 & 2.3 $\pm$ 0.2 \\
         & 1 & & 9.0 $\pm$ 0.0 & 0.4 $\pm$ 0.0 & 0.4 $\pm$ 0.0 & 80.8 $\pm$ 0.3 & 78.7 $\pm$ 0.2 & 2.1 $\pm$ 0.2 \\
         & 5 & & 9.0 $\pm$ 0.0 & 0.2 $\pm$ 0.0 & 0.4 $\pm$ 0.0 & 80.4 $\pm$ 0.2 & 78.1 $\pm$ 0.2 & 2.2 $\pm$ 0.1 \\
         & 10 & & 9.0 $\pm$ 0.6 & 0.2 $\pm$ 0.0 & 0.3 $\pm$ 0.0 & 79.0 $\pm$ 0.2 & 76.6 $\pm$ 0.1 & 2.4 $\pm$ 0.1 \\
         \bottomrule
    \end{tabular}}
    \caption{Metrics on the spatial-alignment test acquired by the baseline TesNet model and TesNet trained with the spatial-aligning training and masking augmentation, using VGG16 backbone instead of Resnet34.}
    \label{tab:results_tesnet_vgg}
\end{table*}

\clearpage
\subsection{Detailed results on Stanford Cars dataset}
In \cref{tab:results_ppnet_cars,tab:results_tesnet_cars}, we show the detailed comparison of spatial alignment metrics between baseline methods and methods improved by the spatial-aligning loss and masking augmentation for ProtoPNet~\cite{chen2019looks} and TesNet~\cite{wang2021interpretable} trained and evaluated on Stanford Cars dataset~\cite{krause2013stanfordcars}. 

\vskip 0.5in

\begin{table*}[t]
    \centering
    \resizebox{\textwidth}{!}{%
    \begin{tabular}{lrc|rrr|ccc}
         \toprule
         method & $\lambda_{align}$ & \multicolumn{1}{c}{MA} & \multicolumn{1}{c}{$\downarrow$ PLC} & \multicolumn{1}{c}{$\downarrow$ PRC} & \multicolumn{1}{c}{$\downarrow$ PAC} & $\uparrow$ Acc. Before & $\uparrow$ Acc. After & $\downarrow$ AC \\
         \toprule
         baseline & \xmark & \xmark & 35.3 $\pm$ 0.0 & 6.4 $\pm$ 0.1 & 28.4 $\pm$ 0.1 & 88.1 $\pm$ 0.0 & 78.1 $\pm$ 0.2 & 10.0 $\pm$ 0.2 \\
         \midrule
         \multirow{7}{*}{ours} & 0.1 & \multirow{3}{*}{\xmark} & 33.0 $\pm$ 2.4 & 5.4 $\pm$ 0.5 & 18.0 $\pm$ 0.8 & 87.9 $\pm$ 0.2 & 77.8 $\pm$ 1.4 & 10.1 $\pm$ 1.2 \\
         & 1 & & 24.3 $\pm$ 2.9 & 12.4 $\pm$ 2.1 & 2.1 $\pm$ 0.7 &88.2 $\pm$ 0.4 & 78.1 $\pm$ 1.0 & 10.2 $\pm$ 0.9\\
         & 5  & & 22.7 $\pm$ 2.8 & 7.5 $\pm$ 1.6 & 1.9 $\pm$ 0.6 & 88.1 $\pm$ 0.4 & 78.1 $\pm$ 1.0 & 10.0 $\pm$ 0.9 \\
         & 10 & & 21.8 $\pm$ 2.5 & 8.4 $\pm$ 1.5 & 1.7 $\pm$ 0.5 & 87.8 $\pm$ 0.2 & 78.5 $\pm$ 1.2 & 10.2 $\pm$ 1.1 \\
         \cmidrule{2-9}
         & 0 & \multirow{4}{*}{\cmark} & 35.2 $\pm$ 1.0 & 6.6 $\pm$ 0.5 & 28.3 $\pm$ 0.7 & 88.1 $\pm$ 0.2 & 77.8 $\pm$ 0.4 & 10.4 $\pm$ 0.4 \\
         & 0.1 & & 25.1 $\pm$ 1.3 & 1.3 $\pm$ 0.3 & 8.1 $\pm$ 0.8 & 88.0 $\pm$ 0.1 & 84.9 $\pm$ 0.4 & 3.0 $\pm$ 0.4 \\
         & 1 & & 15.6 $\pm$ 0.5 & 3.0 $\pm$ 0.2 & 1.6 $\pm$ 0,6 & 88.0 $\pm$ 0.1 & 85.2 $\pm$ 0.2 & 2.8 $\pm$ 0.3 \\
         & 5  & & 13.5 $\pm$ 0.8 & 2.9 $\pm$ 0.2 & 1.4 $\pm$ 0.6 & 88.1 $\pm$ 0.1 & 86.1 $\pm$ 0.3 & 2.0 $\pm$ 0.3 \\
         & 10 & & 12.5 $\pm$ 0.9 & 1.2 $\pm$ 0.3 & 1.3 $\pm$ 0.7 & 88.0 $\pm$ 0.1 & 86.4 $\pm$ 0.2 & 1.4 $\pm$ 0.2 \\
         \bottomrule
    \end{tabular}}
    \caption{Metrics on the spatial-alignment test acquired by the baseline ProtoPNet model and ProtoPNet trained with the spatial-aligning training and masking augmentation, trained and evaluated on the Stanford Cars dataset.}
    \label{tab:results_ppnet_cars}
\end{table*}

\begin{table*}[t]
    \centering
    \resizebox{\textwidth}{!}{%
    \begin{tabular}{lrc|rrr|ccc}
         \toprule
         method & $\lambda_{align}$ & \multicolumn{1}{c}{MA} & \multicolumn{1}{c}{$\downarrow$ PLC} & \multicolumn{1}{c}{$\downarrow$ PRC} & \multicolumn{1}{c}{$\downarrow$ PAC} & $\uparrow$ Acc. Before & $\uparrow$ Acc. After & $\downarrow$ AC \\
         \toprule
         baseline & \xmark & \xmark & 15.3 $\pm$ 0.3 & 1.7 $\pm$ 0.3 & 1.8 $\pm$ 0.2 & 91.8 $\pm$ 0.0 & 86.6 $\pm$ 0.2 & 5.2 $\pm$ 0.2 \\
         \midrule
         \multirow{9}{*}{ours} & 0.1 & \multirow{4}{*}{\xmark} & 15.3 $\pm$ 0.7 & 1.1 $\pm$ 0.1 & 1.1 $\pm$ 0.3 & 91.3 $\pm$ 0.1 & 86.1 $\pm$ 0.2 & 5.2 $\pm$ 0.3 \\
         & 1 & & 13.7 $\pm$ 0.3 & 0.8 $\pm$ 0.1 & 2.7 $\pm$ 0.3 & 91.2 $\pm$ 0.1 & 86.7 $\pm$ 0.1 & 4.5 $\pm$ 0.1 \\
         & 5 & & 16.7 $\pm$ 0.3 & 1.1 $\pm$ 0.1 & 3.0 $\pm$ 0.2 & 89.8 $\pm$ 0.2 & 84.6 $\pm$ 0.4 & 5.2 $\pm$ 0.3 \\
         & 10 & & 15.0 $\pm$ 0.0 & 0.7 $\pm$ 0.1 & 1.3 $\pm$ 0.2 & 89.4 $\pm$ 0.1 & 85.1 $\pm$ 0.1 & 4.3 $\pm$ 0.0 \\
         \cmidrule{2-9}
         & 0 & \multirow{5}{*}{\cmark} & 9.7 $\pm$ 0.3 & 0.5 $\pm$ 0.1 & 0.1 $\pm$ 0.1 & 92.2 $\pm$ 0.1 & 90.9 $\pm$ 0.1 & 1.3 $\pm$ 0.1 \\
         & 0.1 & & 10.0 $\pm$ 0.0 & 0.4 $\pm$ 0.0 & 0.1 $\pm$ 0.1 & 91.9 $\pm$ 0.1 & 90.1 $\pm$ 0.2 & 1.8 $\pm$ 0.2 \\
         & 1 & & 9.0 $\pm$ 0.0 & 0.3 $\pm$ 0.0 & 0.4 $\pm$ 0.1 & 91.7 $\pm$ 0.1 & 90.4 $\pm$ 0.2 & 1.3 $\pm$ 0.2 \\
         & 5 & & 10.3 $\pm$ 0.3 & 0.3 $\pm$ 0.0 & 0.6 $\pm$ 0.0 & 90.7 $\pm$ 0.1 & 89.5 $\pm$ 0.1 & 1.2 $\pm$ 0.1 \\
         & 10 & & 10.7 $\pm$ 0.3 & 0.2 $\pm$ 0.0 & 0.4 $\pm$ 0.0 & 90.1 $\pm$ 0.3 & 88.6 $\pm$ 0.3 & 1.5 $\pm$ 0.1 \\
         \bottomrule
    \end{tabular}}
    \caption{Metrics on the spatial-alignment test acquired by the baseline TesNet model and TesNet trained with the spatial-aligning training and masking augmentation, trained and evaluated on the Stanford Cars dataset.}
    \label{tab:results_tesnet_cars}
\end{table*}

\clearpage

\subsection{Why is TesNet so robust?}

In \cref{tab:results_tesnet_no_losses} shown are results of the benchmark applied to TesNet trained without its specific loss terms (as described in the main text).

\vskip 0.5in

\begin{table*}[t]
    \centering
    \resizebox{\textwidth}{!}{%
    \begin{tabular}{lrc|rrr|ccc}
         \toprule
         method & $\lambda_{align}$ & \multicolumn{1}{c}{MA} & \multicolumn{1}{c}{$\downarrow$ PLC } & \multicolumn{1}{c}{$\downarrow$ PRC} & \multicolumn{1}{c}{$\downarrow$ PAC } & $\uparrow$ Acc. Before & $\uparrow$ Acc. After & $\downarrow$ AC \\
         \toprule
         baseline & \xmark & \xmark & 16.0 $\pm$ 0.0 & 2.9 $\pm$ 0.3 & 3.4 $\pm$ 0.3 & 81.6 $\pm$ 0.2 & 75.8 $\pm$ 0.5 & 5.8 $\pm$ 0.6 \\
         baseline w/o $\lambda_{orth}$ & \xmark & \xmark & 16.0 $\pm$ 0.0 & 3.7 $\pm$ 0.6 & 2.2 $\pm$ 0.1 & 80.0 $\pm$ 0.3 & 73.7 $\pm$ 0.1 & 6.3 $\pm$ 0.2 \\
         baseline w/o $\lambda_{orth}$, $\lambda_{ss}$ & \xmark & \xmark & 15.0 $\pm$ 0.6 & 2.2 $\pm$ 0.5 & 1.6 $\pm$ 0.1 & 80.6 $\pm$ 0.3 & 74.6 $\pm$ 0.1 & 6.0 $\pm$ 0.3 \\
         \bottomrule
    \end{tabular}}
    \caption{Metrics on the spatial-alignment test acquired by the TesNet trained without the TesNet-specific losses.}
    \label{tab:results_tesnet_no_losses}
\end{table*}

\subsection{Results of spatial-aligning training with a smooth mask}
In \cref{tab:results_tesnet_smooth,tab:results_prototree_smooth} we provide results of some additional experiments, obtained for a variant of spatial-aligning training with a smooth mask. In the main series of experiments with the $\lambda_{align}$ loss (described above and in the main text), in the seconds pass we used a binarized mask modifying the original image, that is, mask that was equal 1 inside the highly activated region $b(x)$, and 0 outside of $b(x)$. In order to examine possible effects of such binarization, we conducted additional experiments, this time with a smooth mask, that is, mask taking values from the continuous range $[0, 1]$, proportional to the values of the prototype similarity map $sim(p, f(x))$. The results for TesNet and ProtoTree models suggest that such a 'blurred' mask during training is generally not effective in spatially aligning the prototypes (and might have a strong negative effect on the network performance, as evidenced by results for ProtoTree), in contrast to the binarized version of the mask, which is effective in aligning the prototypes.

\vskip 0.5in

\begin{table*}[t]
    \centering
    \resizebox{\textwidth}{!}{%
    \begin{tabular}{lrc|rrr|ccc}
         \toprule
         method & $\lambda_{align}$ & \multicolumn{1}{c}{MA} & \multicolumn{1}{c}{$\downarrow$ PLC } & \multicolumn{1}{c}{$\downarrow$ PRC} & \multicolumn{1}{c}{$\downarrow$ PAC } & $\uparrow$ Acc. Before & $\uparrow$ Acc. After & $\downarrow$ AC \\
         \toprule
         baseline & \xmark & \xmark & 16.0 $\pm$ 0.0 & 2.9 $\pm$ 0.3 & 3.4 $\pm$ 0.3 & 81.6 $\pm$ 0.2 & 75.8 $\pm$ 0.5 & 5.8 $\pm$ 0.6 \\
         \midrule
         \multirow{7}{*}{ours} & 0.1 & \multirow{4}{*}{\xmark} & 17.7 $\pm$ 0.3 & 3.4 $\pm$ 0.1 & 4.1 $\pm$ 0.4 & 81.5 $\pm$ 0.2 & 74.0 $\pm$ 0.2 & 7.5 $\pm$ 0.2 \\
         &  1 & & 16.0 $\pm$ 0.0 & 3.1 $\pm$ 0.2 & 3.9 $\pm$ 0.2 & 81.3 $\pm$ 0.2 & 74.9 $\pm$ 0.3 & 6.4 $\pm$ 0.1 \\
         & 10 & & 15.0 $\pm$ 0.6 & 1.9 $\pm$ 0.3 & 5.1 $\pm$ 0.4 & 80.8 $\pm$ 0.1 & 75.1 $\pm$ 0.3 & 5.7 $\pm$ 0.2 \\
         \cmidrule{2-9}
         & 0 & \multirow{4}{*}{\cmark} & 9.0 $\pm$ 0.0 & 0.4 $\pm$ 0.0 & 0.1 $\pm$ 0.0 & 82.6 $\pm$ 0.2 & 81.5 $\pm$ 0.2 & 1.9 $\pm$ 0.1 \\
         & 0.1 & & 9.7 $\pm$ 0.3 & 0.5 $\pm$ 0.0 & 0.1 $\pm$ 0.0 & 82.2 $\pm$ 0.1 & 80.3 $\pm$ 0.2 & 1.9 $\pm$ 0.1 \\
         & 1 & & 10.0 $\pm$ 0.0 & 0.5 $\pm$ 0.0 & 0.3 $\pm$ 0.1 & 82.9 $\pm$ 0.1 & 80.6 $\pm$ 0.3 & 2.4 $\pm$ 0.2 \\
         & 10 & & 14.0 $\pm$ 4.0 & 0.7 $\pm$ 0.4 & 0.6 $\pm$ 0.2 & 81.2 $\pm$ 0.7 & 79.0 $\pm$ 0.7 & 2.3 $\pm$ 0.1 \\
         \bottomrule
    \end{tabular}}
    \caption{Metrics on the spatial-alignment test acquired by the baseline TesNet model (included here for ease of reference) and TesNet trained with the spatial-aligning training with a smooth mask and masking augmentation.}
    \label{tab:results_tesnet_smooth}
\end{table*}

\clearpage

\begin{table*}[t]
    \centering
    \resizebox{\textwidth}{!}{%
    \begin{tabular}{lrc|rrr|ccc}
         \toprule
         method & $\lambda_{align}$ & \multicolumn{1}{c}{MA} & \multicolumn{1}{c}{$\downarrow$ PLC } & \multicolumn{1}{c}{$\downarrow$ PRC} & \multicolumn{1}{c}{$\downarrow$ PAC } & $\uparrow$ Acc. Before & $\uparrow$ Acc. After & $\downarrow$ AC \\
         \toprule
         baseline & \xmark & \xmark & 27.7 $\pm$ 0.3 & 12.7 $\pm$ 1.3 & 6.5 $\pm$ 0.1 & 82.5 $\pm$ 0.2 & 70.4 $\pm$ 1.2 & 12.1 $\pm$ 1.0 \\
         \midrule
         \multirow{7}{*}{ours} & 0.1 & \multirow{4}{*}{\xmark} & 31.3 $\pm$ 0.3 & 19.6 $\pm$ 2.5 & 10.7 $\pm$ 0.3 & 79.4 $\pm$ 1.3 & 64.3 $\pm$ 1.9 & 15.1 $\pm$ 0.6 \\
         &  1 & & 30.3 $\pm$ 1.9 & 18.8 $\pm$ 4.9 & 8.9 $\pm$ 1.6 & 68.6 $\pm$ 7.5 & 59.8 $\pm$ 3.1 & 8.9 $\pm$ 4.5 \\
         & 10 & & 31.7 $\pm$ 1.7 & 19.5 $\pm$ 1.1 & 8.0 $\pm$ 2.3 & 78.9 $\pm$ 0.7 & 60.6 $\pm$ 3.4 & 18.3 $\pm$ 2.8 \\
         \cmidrule{2-9}
         & 0 & \multirow{4}{*}{\cmark} & 21.7 $\pm$ 0.3 & 8.4 $\pm$ 0.5 & 2.5 $\pm$ 0.4 & 82.9 $\pm$ 0.5 & 78.0 $\pm$ 0.6 & 4.9 $\pm$ 0.2 \\
         & 0.1 & & 23.0 $\pm$ 0.6 & 6.8 $\pm$ 0.6 & 3.3 $\pm$ 0.7 & 78.3 $\pm$ 2.5 & 73.7 $\pm$ 2.2 & 4.6 $\pm$ 0.4 \\
         & 1 & & 22.7 $\pm$ 0.9 & 12.1 $\pm$ 2.2 & 4.7 $\pm$ 1.5 & 71.2 $\pm$ 4.8 & 68.1 $\pm$ 3.5 & 3.1 $\pm$ 1.3 \\
         & 10 & & 24.7 $\pm$ 0.7 & 10.1 $\pm$ 0.3 & 3.6 $\pm$ 0.7 & 79.8 $\pm$ 0.3 & 74.6 $\pm$ 0.6 & 5.2 $\pm$ 0.9 \\
         \bottomrule
    \end{tabular}}
    \caption{Metrics on the spatial-alignment test acquired by the baseline ProtoTree model (included here for ease of reference) and ProtoTree trained with the spatial-aligning training with a smooth mask and masking augmentation. }
    \label{tab:results_prototree_smooth}
\end{table*}

\end{document}